%% file: paper.tex
\documentclass[format=acmsmall, review=false, screen=true]{acmart}

\input{hri-macros}

\usepackage{booktabs} % For formal tables
\usepackage[subpreambles=true]{standalone}

\acmJournal{THRI} %UNSURE

\usepackage{multicol}
\usepackage[font = footnotesize]{subcaption}

\usepackage{array}
\usepackage{booktabs}
\usepackage{tabularx}
\usepackage{multirow}

\usepackage{soul}
\usepackage{amsmath}
\usepackage{tikz}
\usepackage{todonotes}
\usepackage{enumitem}
\usepackage{flushend}

\usepackage{xspace}
\usepackage{subcaption}

\usepackage{graphicx}
\graphicspath{{figs/}}
    \DeclareGraphicsExtensions{.pdf,.jpg,.png,.eps}
\usepackage{amsfonts}
\usepackage{float}
\usepackage{enumerate}
\usepackage{fancybox}
\usepackage{color}

\usetikzlibrary{arrows,decorations.markings}

\newcommand{\ie}{\textit{i.e.}}
\newcommand{\eg}{\textit{e.g.}}

\newcommand{\secref}[1]{Section~\ref{#1}}

\newcommand{\equref}[1]{Eq.~\ref{#1}}

\newcommand{\figref}[1]{Figure~\ref{#1}}

\newcommand{\tabref}[1]{Table~\ref{#1}}

\DeclareMathOperator*{\argmax}{arg\,max}

\newcommand{\xxnote}[3]{}
\ifx\hidenotes\undefined
  \usepackage{color}
  \renewcommand{\xxnote}[3]{\color{#2}{#1: #3}}
\fi

\newcommand{\textrmsmall}[1]{\textrm{\fontsize{6}{7.2}\selectfont{#1}}}

\title{Trust-Aware Decision Making for Human-Robot Collaboration: Model Learning and Planning}

\author{Min Chen}
\authornote{Chen and Nikolaidis contributed equally to the work.}
\affiliation{%
    \institution{National University of Singapore}
}
\email{chenmin@comp.nus.edu.sg}
\author{Stefanos Nikolaidis}
\affiliation{%
    \institution{University of Southern California}
}
\email{snikolai@alumni.cmu.edu}

\author{Harold Soh}
\affiliation{%
    \institution{National University of Singapore}
}
\email{harold@comp.nus.edu.sg}

\author{David Hsu}
\affiliation{%
    \institution{National University of Singapore}
}
\email{dyhsu@comp.nus.edu.sg}

\author{Siddhartha Srinivasa}
\affiliation{%
    \institution{University of Washington}
}
\email{siddh@cs.uw.edu}

\renewcommand{\shortauthors}{M. Chen et al.}

\begin{document}

\begin{abstract}

Trust in autonomy is essential for effective human-robot collaboration and user adoption of autonomous systems such as robot assistants. This paper introduces  a computational model which integrates trust into robot decision-making. Specifically, we learn from data a partially observable Markov decision process (POMDP) with human trust as a latent variable. The trust-POMDP model provides a principled approach for the robot to (i) infer the trust of a human teammate through interaction, (ii) reason about the effect of its own actions on human trust, and (iii) choose actions that maximize team performance over the long term.   We validated the model through human subject experiments on a table-clearing task in simulation (201 participants) and with a real robot (20 participants). In our studies, the robot builds human trust by manipulating low-risk objects first. Interestingly, the robot sometimes fails intentionally in order to modulate human trust and achieve the best team performance.  These results show that the trust-POMDP calibrates trust to improve human-robot team performance over the long term.  Further, they highlight that maximizing trust alone does not always lead to the best performance. 

\end{abstract}

\begin{CCSXML}
    <ccs2012>
    <concept>
    <concept_id>10003120.10003121.10003124.10011751</concept_id>
    <concept_desc>Human-centered computing~Collaborative interaction</concept_desc>
    <concept_significance>500</concept_significance>
    </concept>
    <concept>
    <concept_id>10010147.10010178.10010199.10010201</concept_id>
    <concept_desc>Computing methodologies~Planning under uncertainty</concept_desc>
    <concept_significance>300</concept_significance>
    </concept>
    </ccs2012>
\end{CCSXML}

\ccsdesc[500]{Human-centered computing~Collaborative interaction}
\ccsdesc[300]{Computing methodologies~Planning under uncertainty}

\keywords{Trust models, Human-robot collaboration, Partially observable Markov decision process (POMDP)}

\maketitle

\section{Introduction}
\label{sec:intro}

%% motivation
%% trust and performance
Trust is essential for seamless human-robot collaboration and user adoption of autonomous systems, such as robot assistants.  Over-trusting robot autonomy may lead to misuse of such systems, where people rely excessively on automation, failing to intervene in the case of critical failures~\citep{lee2004trust}. On the other hand, lack of trust leads to disuse of autonomous systems: users ignore the systems' capabilities, with negative effects on overall performance. 

\begin{figure}[t]
\centering
% \begin{subfigure}[b]{0.85\linewidth}
% \centering
%\includegraphics[width=\linewidth]{figs/table_clearing_task_1.jpg}
\includegraphics[width=0.8\linewidth]{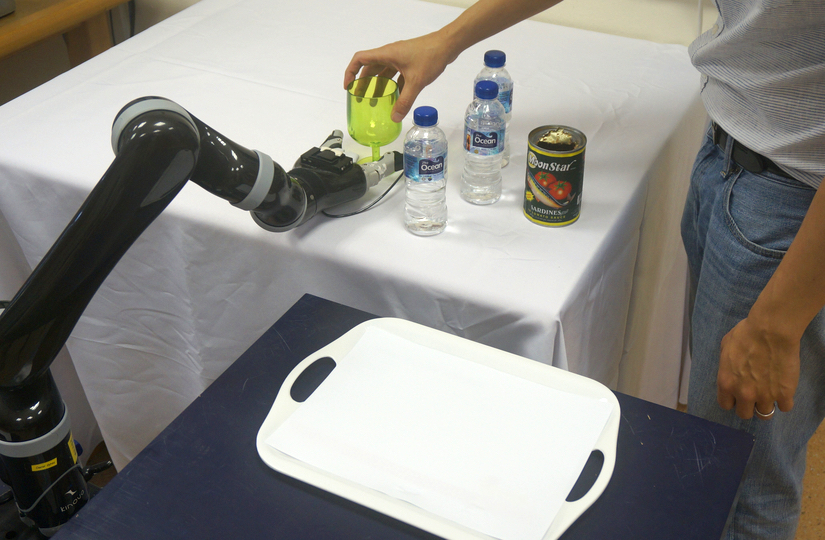}
\caption{
A robot and a human collaborate to clear a table. The human, with low initial trust in the robot, intervenes to stop the robot from moving the wine glass.
}
\label{fig:front}
\end{figure}

We witnessed an example of users' distrust in the system in one of our studies, where a human participant and a robot collaborated to clear a table (\figref{fig:front}). %A user that has a high trust in the robot is likely to let the robot pick up the glass cup and move it to the tray. 
Although the robot was fully capable of handling all objects on the table,  inexperienced participants did not trust that the robot was able to succeed and stopped the robot from moving the wine glass, since they were afraid that the glass may fall and break. It was clear that their trust was poorly calibrated with respect to the robot's true capabilities. This, in turn, had a significant effect on the interaction. 

%% trust-POMDP
%Clearly human trust in the robot directly affected the perception of risk~\citep{siegrist2000influence} and consequently, the interaction. 

This study revealed that, in order to achieve fluent human-robot collaboration, the robot should \emph{monitor} human trust and \emph{influence} it so that it matches the system's capabilities. In our study, for instance, the robot should build human trust first by acting in a trustworthy manner, before going for the wine glass. 

We propose a trust-based computational model of robot decision making: Since trust is not fully observable, we model it as a latent variable in a partially observable Markov decision process (POMDP)~\citep{KaeLit98}. Our trust-POMDP model contains two key components: (i) a trust dynamics model, which captures the evolution of human trust in the robot, and (ii) a human decision model, which connects trust with human actions. Our POMDP formulation can accommodate a variety of trust dynamics and human decision models. Here, we adopt a data-driven approach and learn these models from data.

\begin{figure*}[t!]
\setlength
\tabcolsep{0pt}
\captionsetup[subfigure]{labelformat=empty}
\captionsetup[subfigure]{width=0.68\columnwidth}
\captionsetup[subfigure]{justification=justified,singlelinecheck=false}

\centering

\begin{tabular}{ccccc|c}
%\begin{subfigure}[l]{.05\linewidth}
%\centering
%\small  \vspace{0.8cm}  Trust-POMDP \small  ~\\ \vspace{2.8cm} Myopic 
%\end{subfigure}
%&
\begin{subfigure}[b]{.13\linewidth}
\centering
  \begin{tabular}{cc}
  \\
      \includegraphics[width=0.9\linewidth]{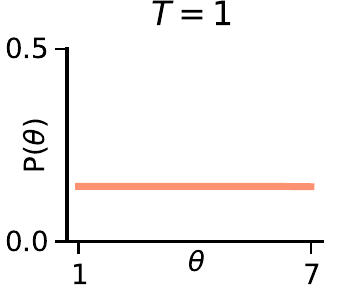}\\
  \includegraphics[width=0.85\linewidth]{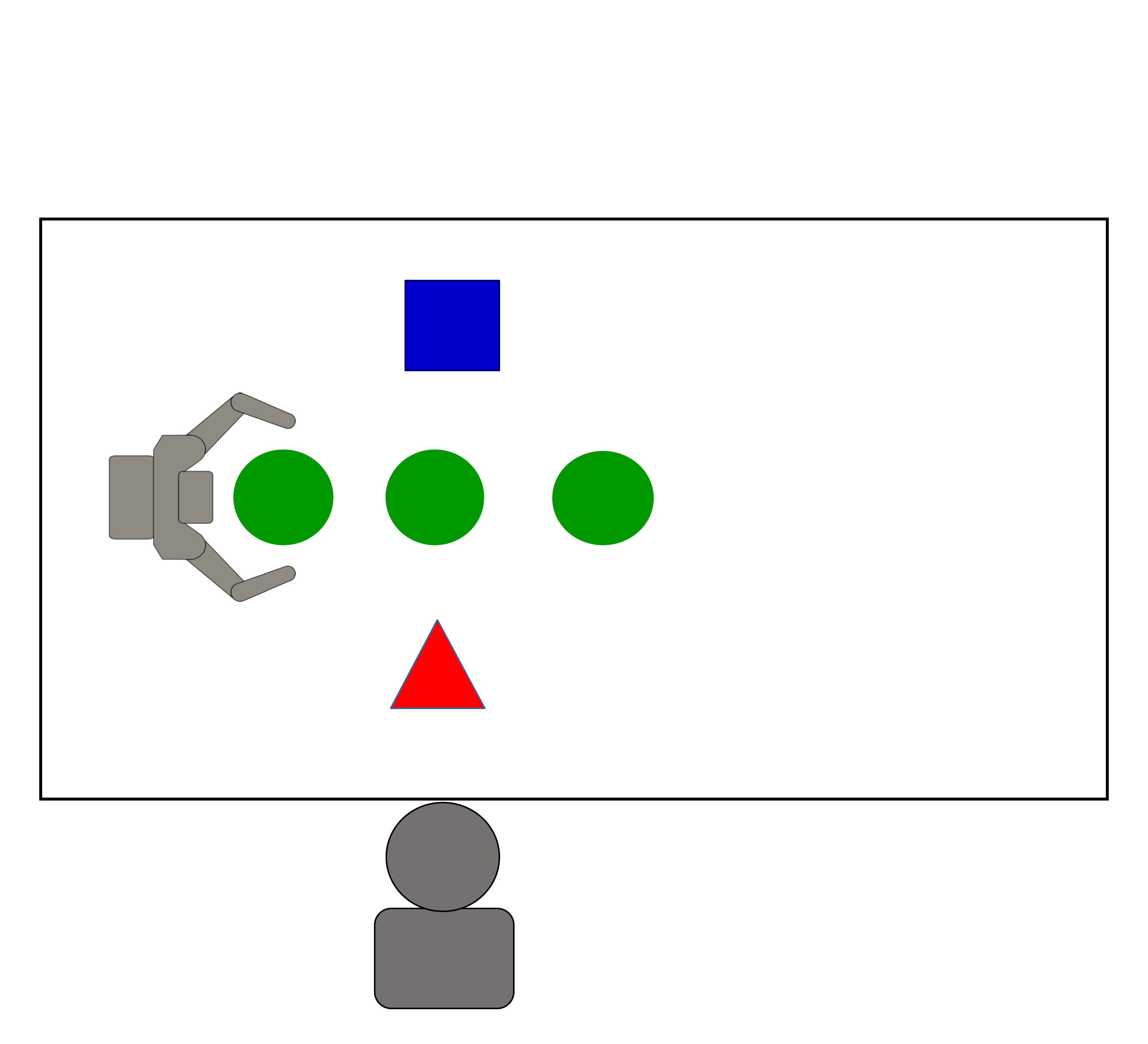} \\
  \\
  \\
 \includegraphics[width=0.85\linewidth]{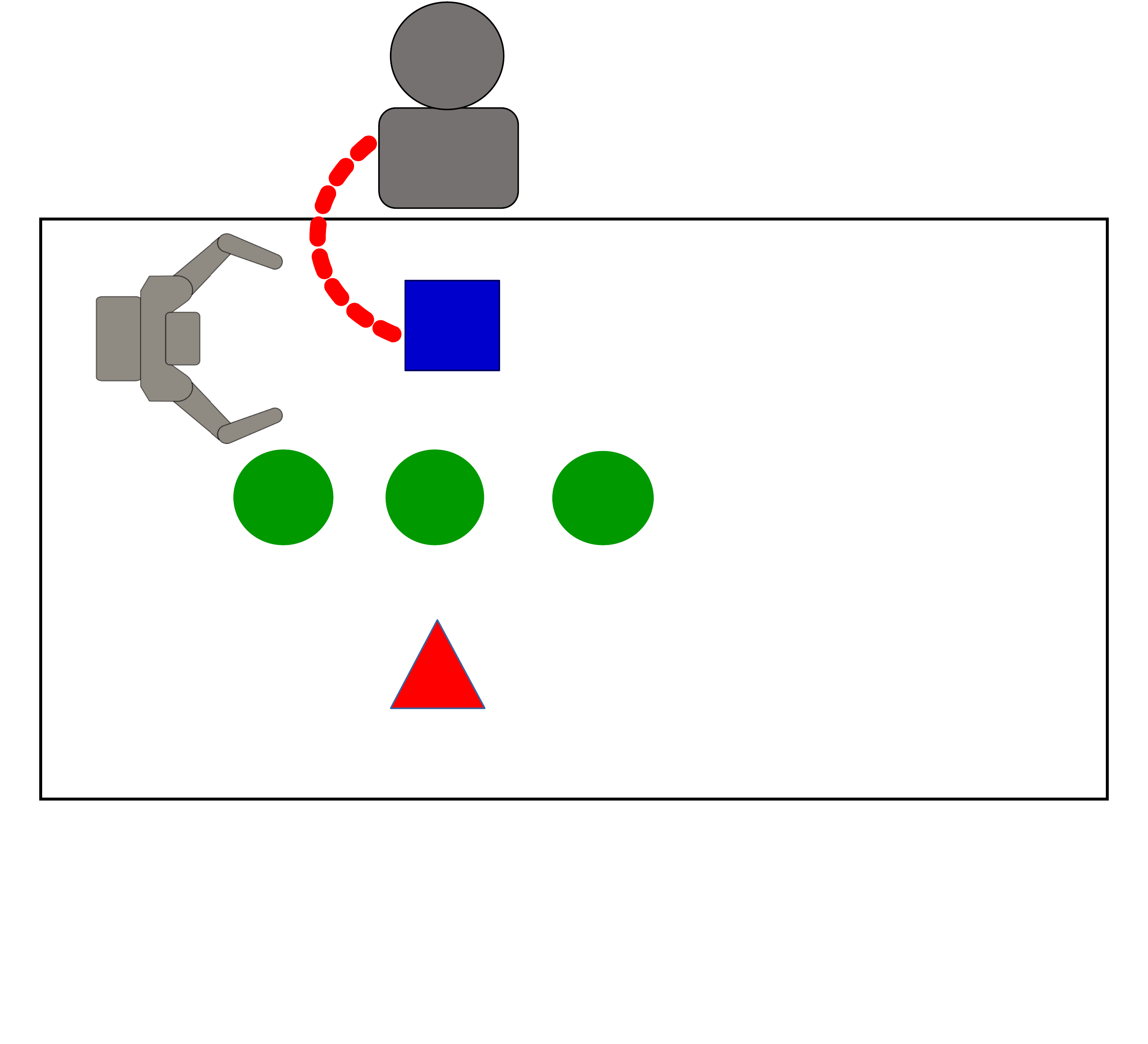} 
   \end{tabular}
   \label{fig:plotT0}
\end{subfigure}
&
\begin{subfigure}[b]{.13\linewidth}
\centering
  \begin{tabular}{cc}
      \\
      \includegraphics[width=0.9\linewidth]{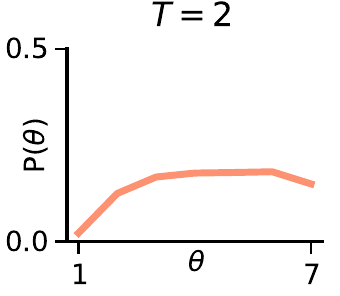}\\
  \includegraphics[width=0.85\linewidth]{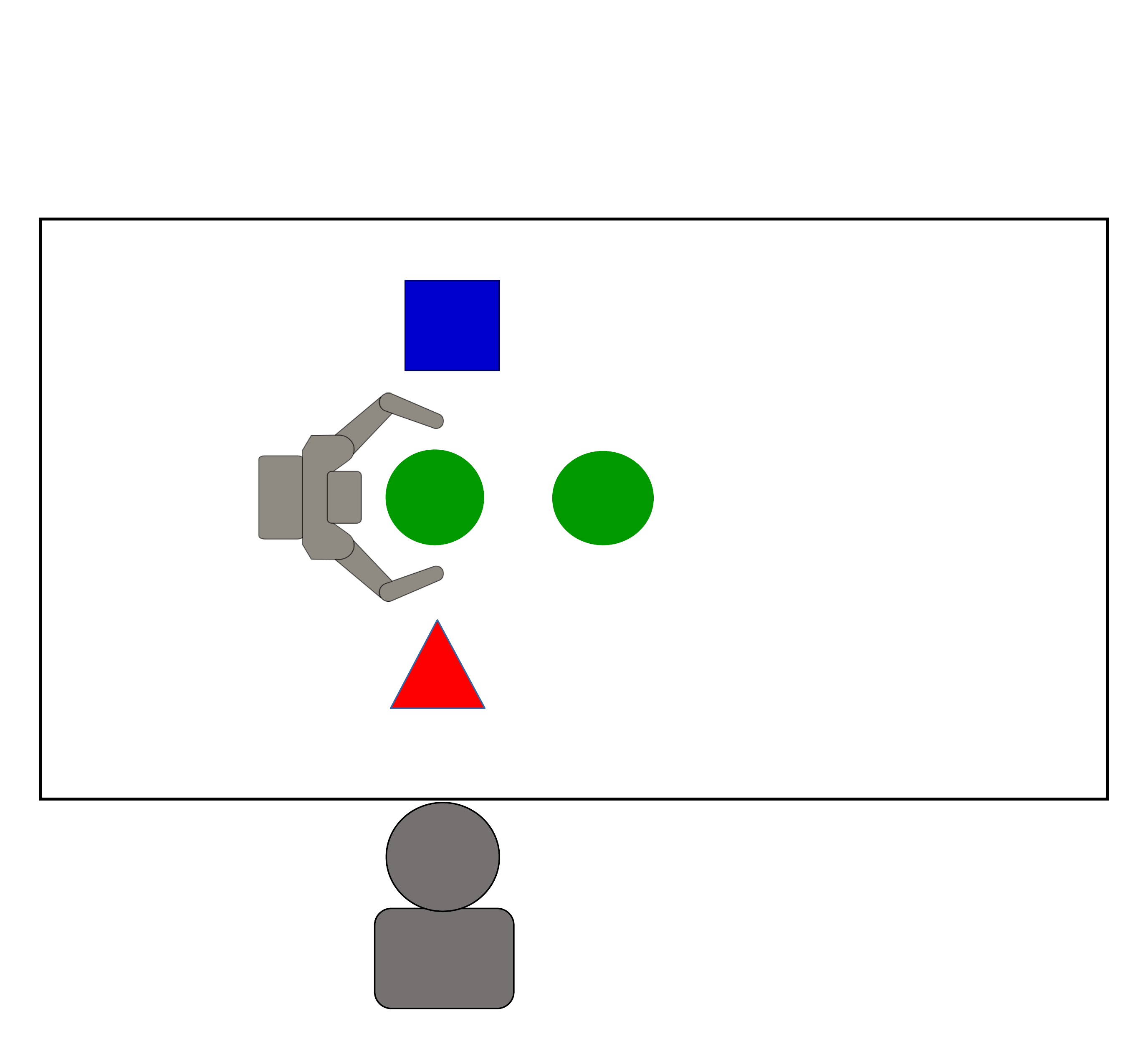} \\
  \\
  \\
 \includegraphics[width=0.85\linewidth]{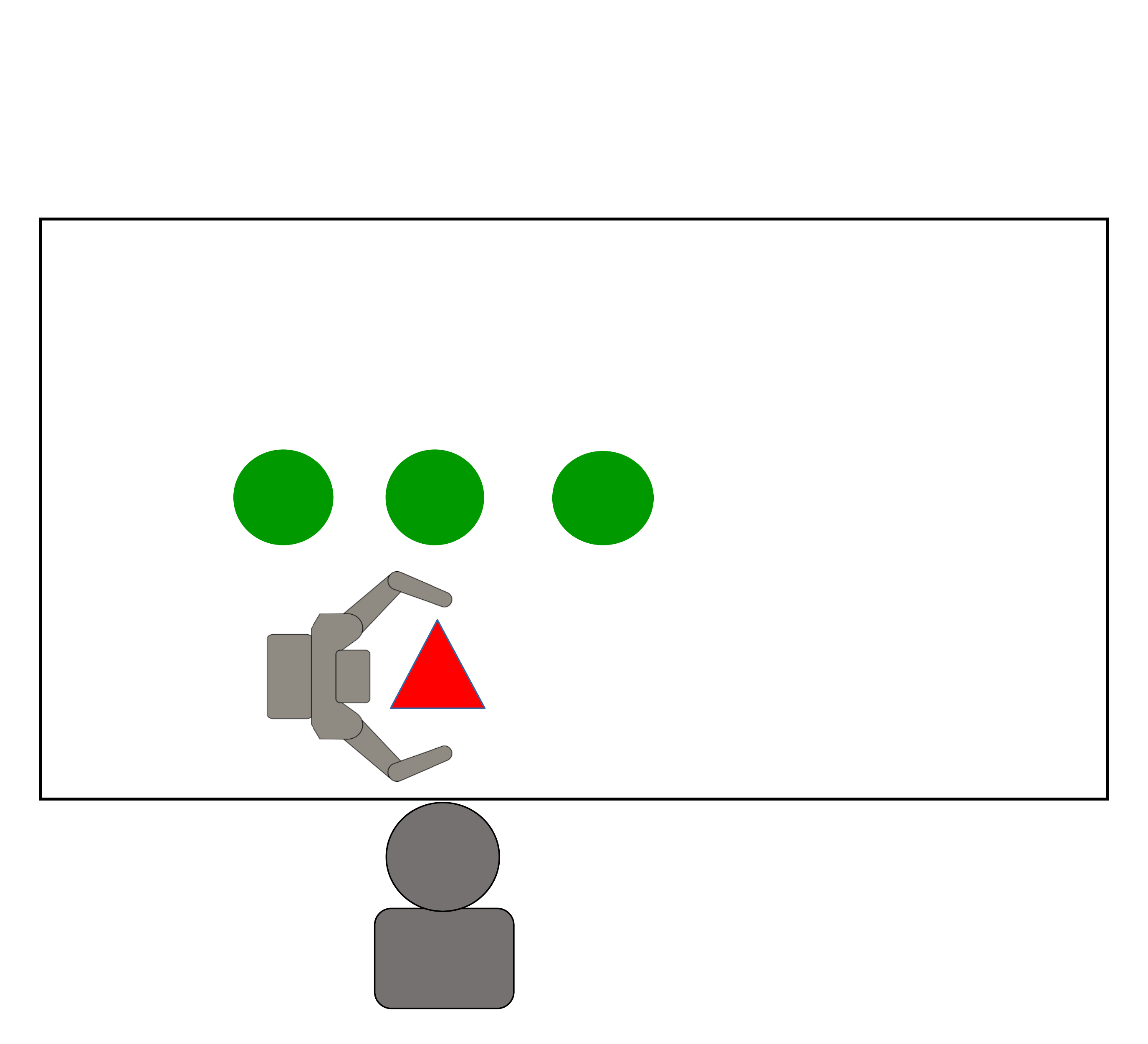} 
   \end{tabular}
   \label{fig:plotT1}
\end{subfigure}
&
\begin{subfigure}[b]{.13\linewidth}
\centering
  \begin{tabular}{cc}
  \fontsize{7}{5}\selectfont
  %\textbf{Trust-POMDP}\par\medskip
  \textbf{Trust-POMDP} \\
  
      \includegraphics[width=0.9\linewidth]{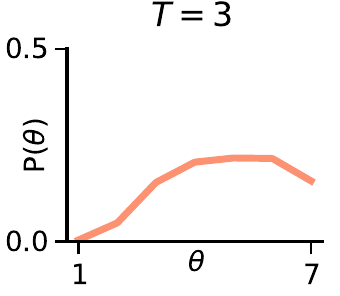}\\
  \includegraphics[width=0.85\linewidth]{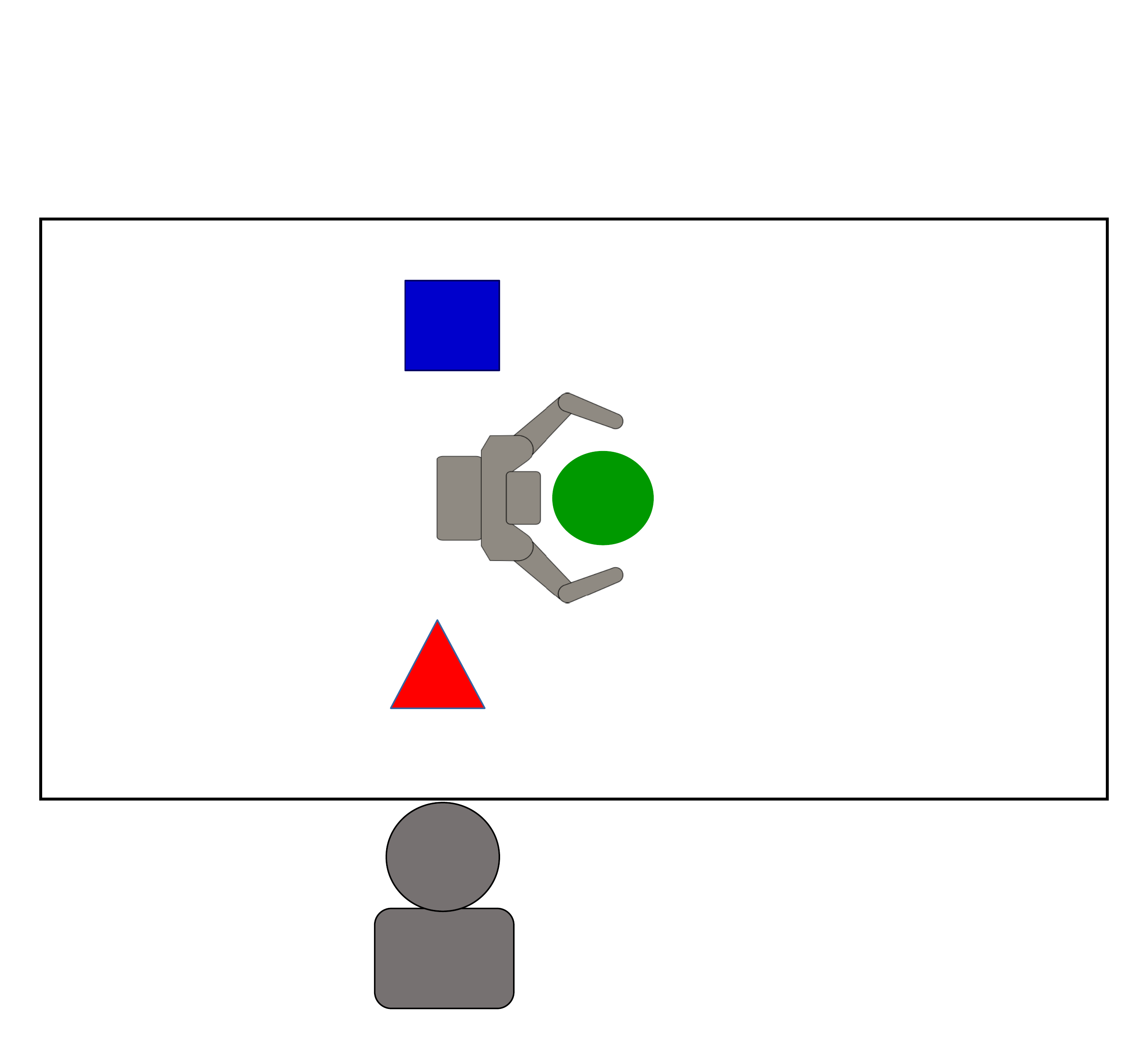} \\
  \\
  \fontsize{7}{5}\selectfont
  %\textbf{\Baseline}\par\medskip
  \textbf{\Baseline}\\
 \includegraphics[width=0.85\linewidth]{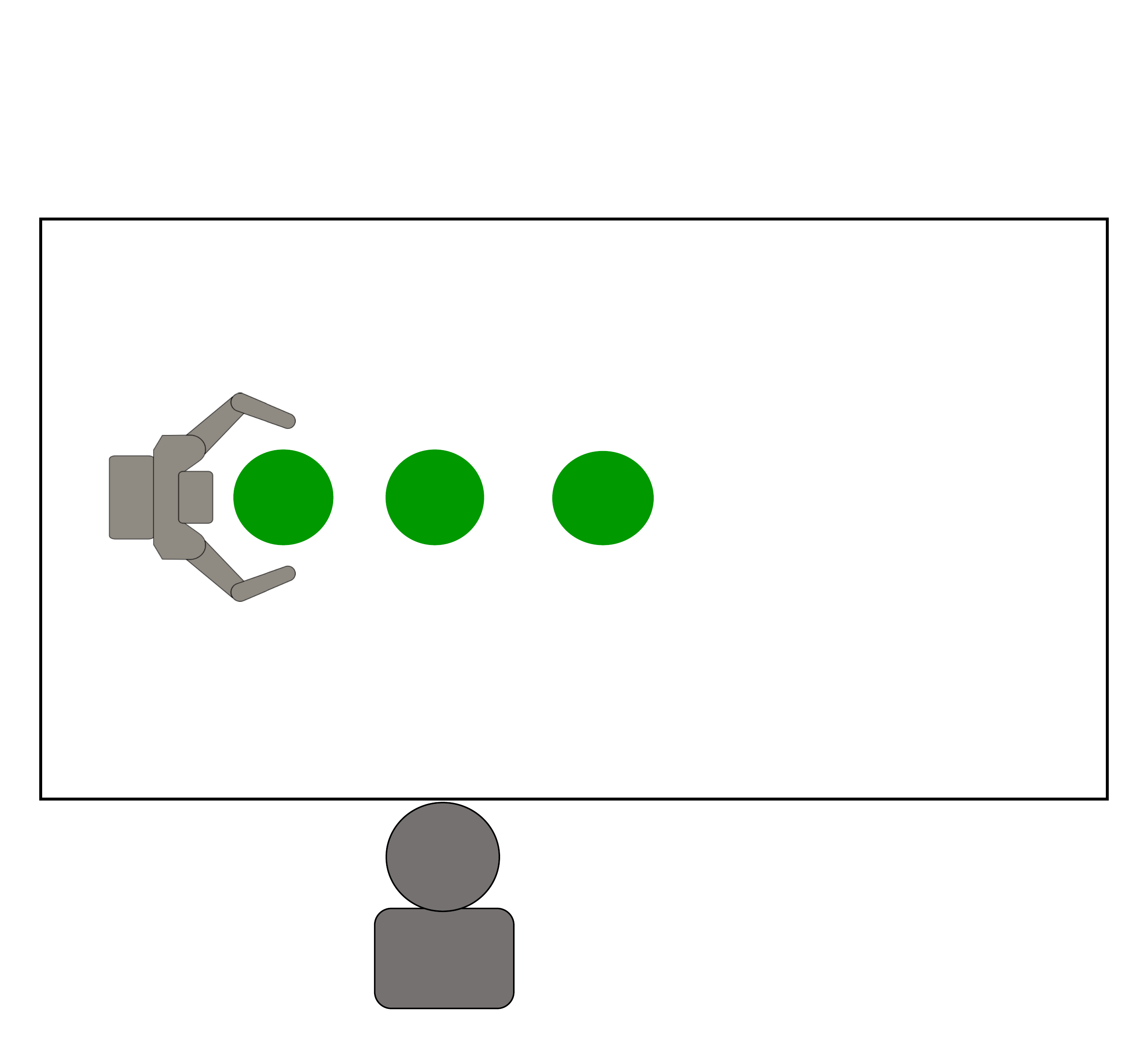} 

   \end{tabular}
   \label{fig:plotT2}
\end{subfigure}
&
\begin{subfigure}[b]{.13\linewidth}
\centering
  \begin{tabular}{cc}
      \\
      \includegraphics[width=0.9\linewidth]{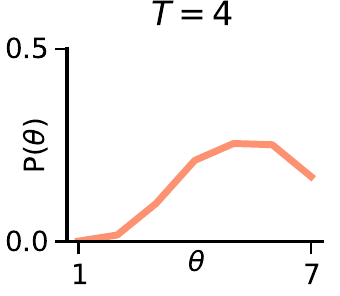}\\
  \includegraphics[width=0.85\linewidth]{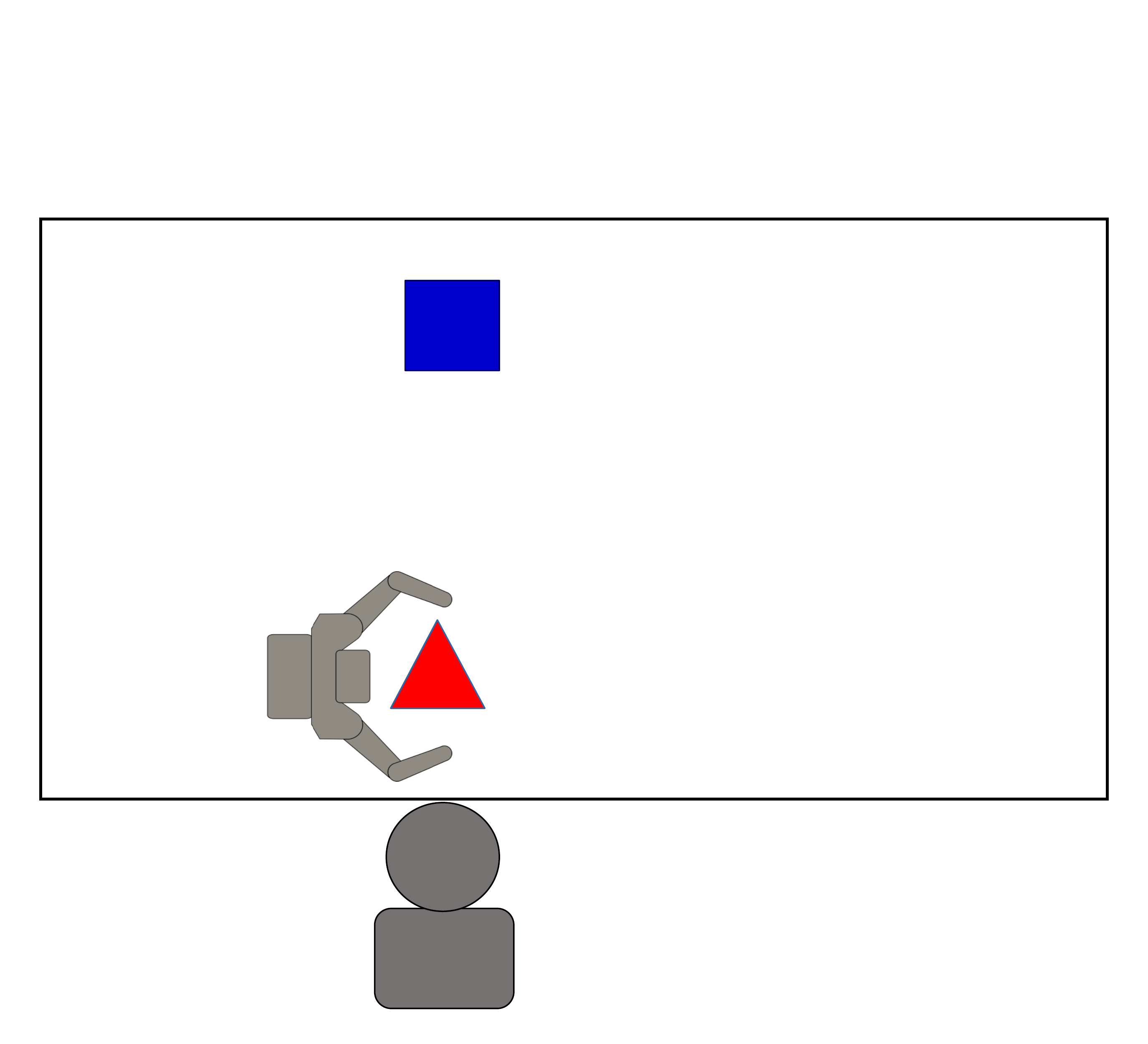} \\
  \\
  \\
 \includegraphics[width=0.85\linewidth]{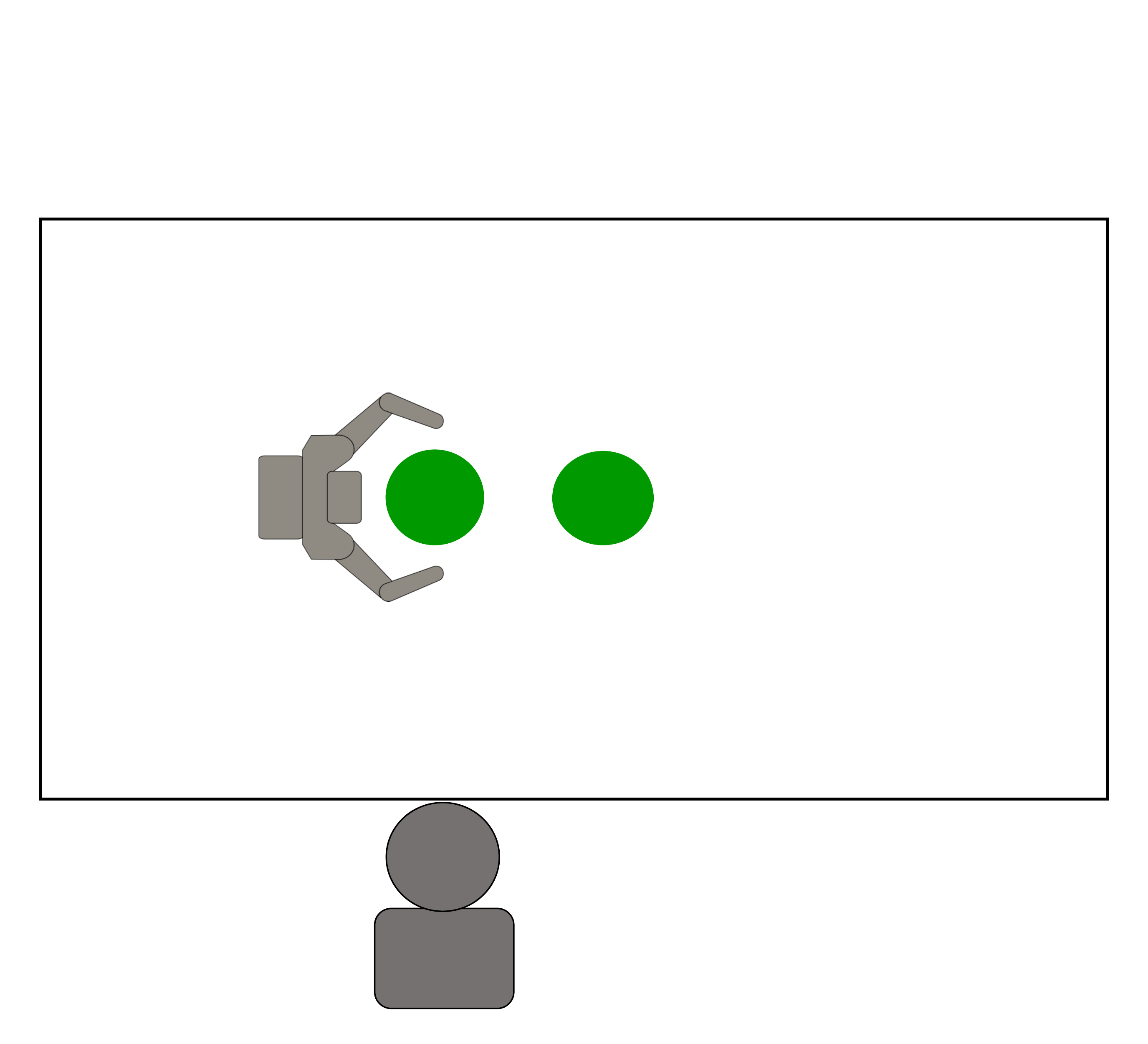} 
   \end{tabular}
   \label{fig:plotT3}
\end{subfigure}
&
\begin{subfigure}[b]{.13\linewidth}
\centering
  \begin{tabular}{cc}
      \\
      \includegraphics[width=0.9\linewidth]{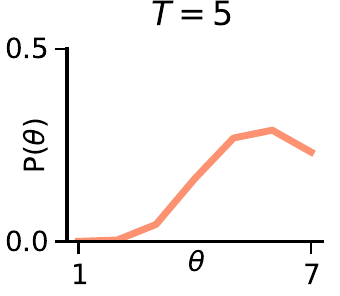}\\
  \includegraphics[width=0.85\linewidth]{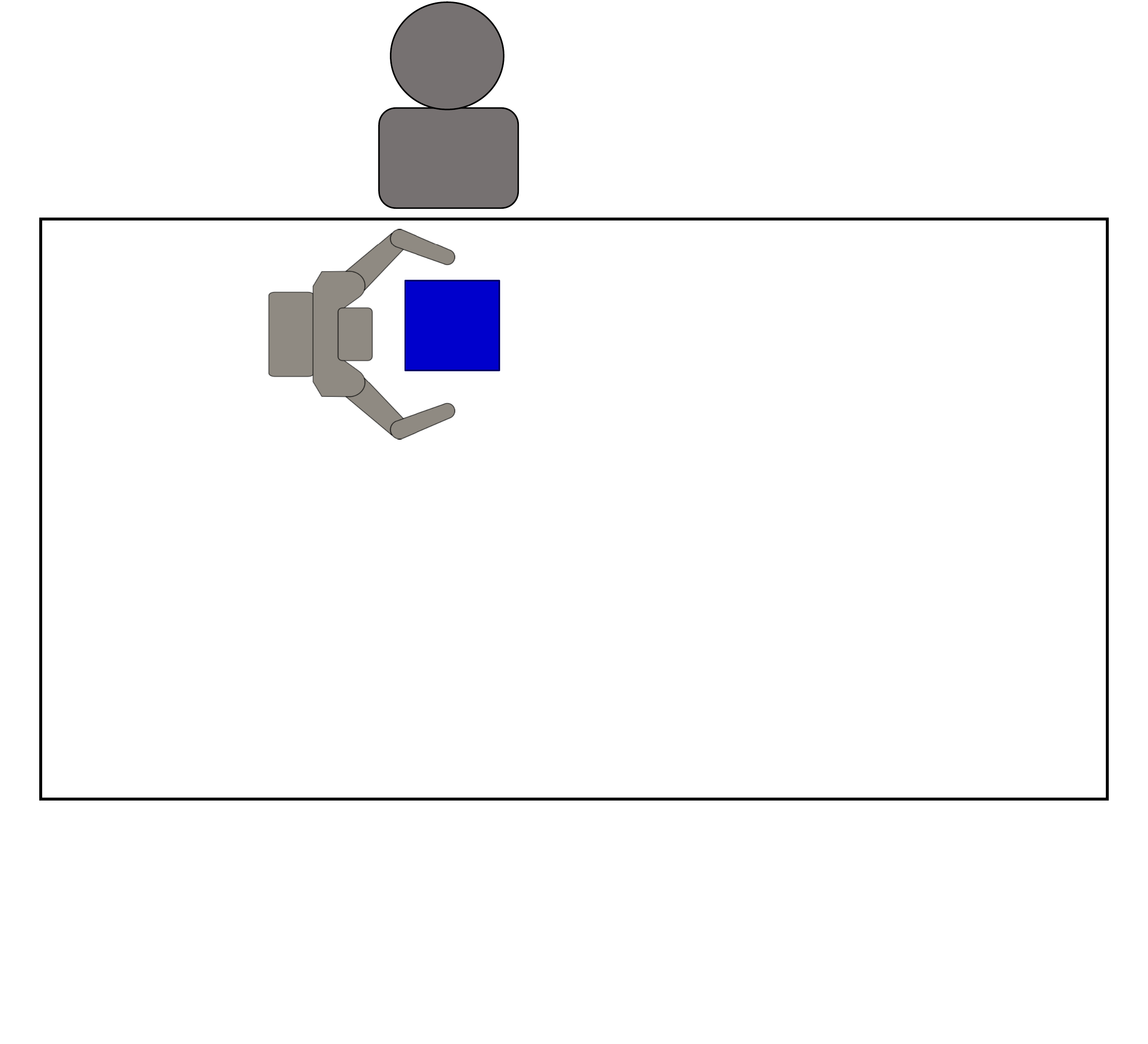} \\
  \\
  \\
 \includegraphics[width=0.85\linewidth]{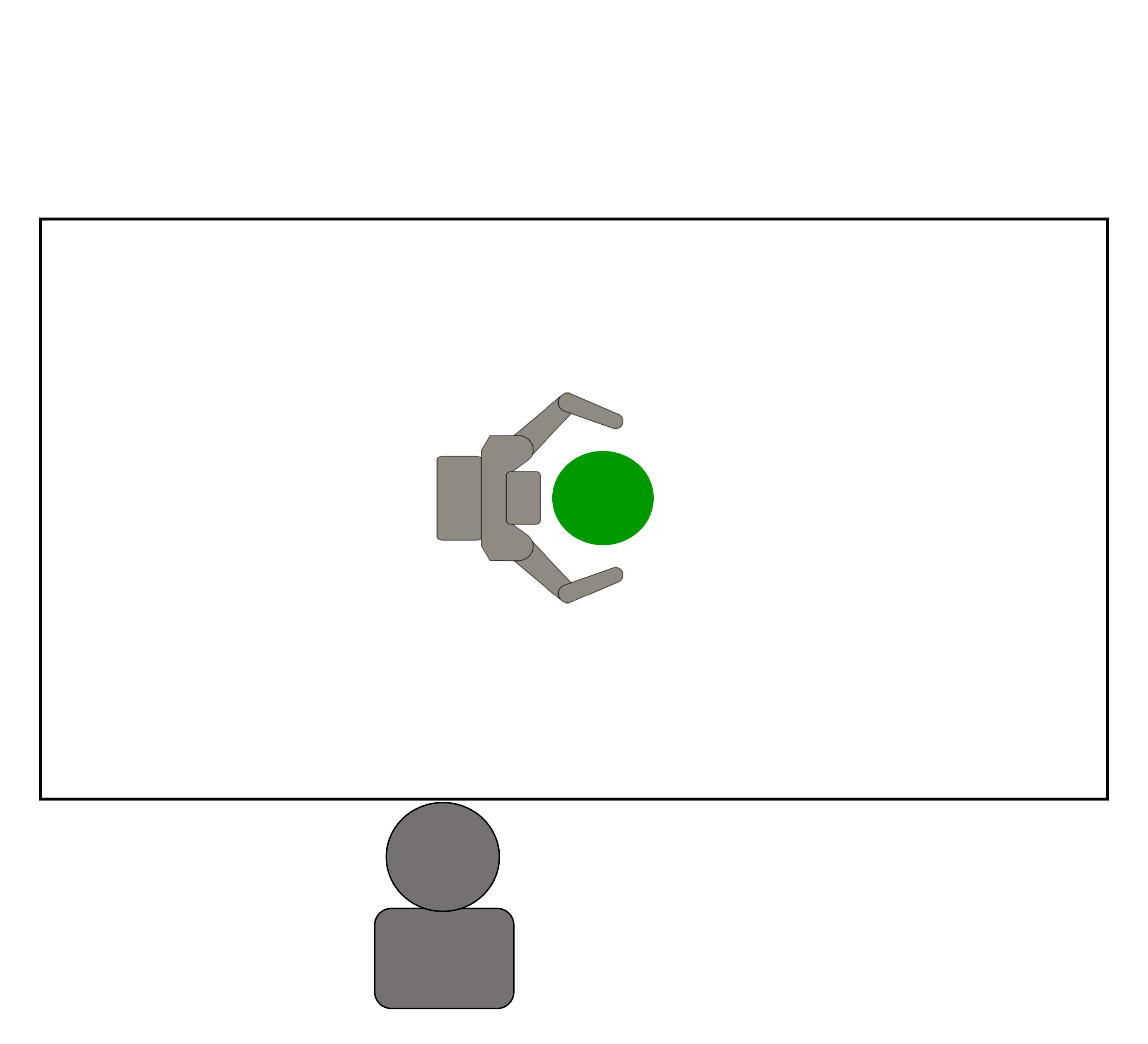} 
   \end{tabular}
   \label{fig:plotT4}
\end{subfigure}
&
\begin{subfigure}[c]{.35\linewidth}
\centering
 \includegraphics[width=0.67\linewidth]{figs/task_images/green_glass_intervene.jpg}\\ 
 ~\\
 \subcaption{
     \mycircle{green} \hskip 14pt Bottle \\
     \mytriangle{red} \hskip 14pt Can\\ 
     \mysquare{blue} \hskip 14pt Glass
     %\\
     %\myarrow  \xspace \xspace \xspace  Human Intervention
 }
 \label{fig:plotT5}
\end{subfigure}
\end{tabular}
\caption{Sample runs of the trust-POMDP strategy and the \baseline strategy on a collaborative table-clearing task.
    The top row shows the probabilistic estimates of human trust over time on a 7-point Likert scale.  
The trust-POMDP strategy starts by moving the plastic  bottles to build trust ($T=1,2,3$) and moves the wine glass only when the estimated trust is high enough ($T=5$). The \baseline strategy does not account for trust and starts with the wine glass, causing the human with low initial trust to intervene ($T=1$).}
\label{fig:behaviors}
\end{figure*}

%% benefits
Although prior work has studied   
human trust elicitation and modeling~\citep{lee1992trust,Floyd2015,xu2015optimo,wang2016trust}, we close the loop between  trust modeling and robot decision-making. The trust-POMDP enables the robot to  systematically infer and influence the human collaborator's trust,  and leverage trust for improved human-robot collaboration and long-term task performance. 

Consider again the table clearing example (\figref{fig:behaviors}). 
 The trust-POMDP strategy first removes the three plastic water bottles to build up trust and only attempts to remove the wine glass  afterwards. In contrast, a baseline myopic strategy  maximizes short-term task performance and does not account for human trust in choosing the robot actions. It first removes the wine glass, which offers the highest reward,  resulting in  unnecessary interventions by human collaborators with low initial trust.

We validated the trust-POMDP model through human subject experiments 
on the collaborative table-clearing task, both online in simulation (201 participants) and with a real robot (20 participants). Compared with the myopic strategy,  the trust-POMDP strategy significantly reduced participants' intervention rate, indicating improved team collaboration and task performance. 

In these experiments the robot always succeeded. Robots, however, fail frequently. What if the robot is likely to \emph{fail} when picking up the wine glass? The robot should then assess human trust in the beginning of the task; if trust is too high, the robot should effectively \emph{communicate} this to the human, in order to calibrate human trust to the appropriate level. While human teammates are able to use natural language to communicate expectations~\cite{mathieu2000influence}, our assistive robotic arm does not have verbal communication capabilities. The trust-POMDP strategy in this case enables the robot to modulate human trust by \emph{intentionally failing} when picking up the bottles, before attempting to grasp the wine glass. This prompts the human to intervene when the robot attempts to pick up the wine glass, preventing failure. 

This paper builds upon our previous work~\cite{chen2018planning} by introducing robot failures into the computational framework. In particular, (i) we augment the dynamics model with robot failures, add a new session of data collection to learn the model and discuss the effect of failures on different levels of trust; (ii) we simulate and visualize robot policies with the learned model; (iii) we provide an analysis of the results in the case of an adaptive policy that enables the robot to assess participants' initial trust and intentionally fail.

% In a broader context, 
Integrating trust modeling and robot decision making enables  robot
behaviors that leverage human trust and actively \emph{modulate} it for
seamless human-robot collaboration.  Under the trust-POMDP
model, the robot deliberately chooses to fail in order to reduce the trust
of an overly trusting user and achieve better task performance over the long
term.  Further, embedding trust in a reward-based POMDP framework makes our
robot task-driven: when the human collaboration is unnecessary, the robot may set aside trust building and act to maximize the team task performance directly.  
All these diverse behaviors emerge automatically from the trust-POMDP model, without explicit manual robot programming.

% We look forward to the resolution of these outstanding issues and to applications of our trust-POMDP to enhance the usability of autonomous systems in a variety of collaborative settings. 

%serves as a precursor to the development of an exciting range of robot behaviors that manipulate trust, in order to mitigate misuse or disuse of autonomous systems. We discuss several limitations of this work and possible extensions in \secref{sec:Discussion}. Key among them are the inclusion of different failure cases in the model, and the testing of settings where our robot reduces the trust of an overly trusting user. 

%Finally, we look forward to applications of these ideas for the development of an exciting range of robot behaviors that manipulate trust, in order to mitigate misuse or disuse of autonomous systems. 
 %performed significantly better with our robot, compared with a baseline robot that did not account for human trust. The difference in performance occurred, because they intervened

%. For instance, Fig.~\ref{fig:behaviors} compares a representative sample run under our proposed trust-POMDP to the baseline policy. The baseline policy started by first removing the highest reward item (the glass cup), which was presumably optimal given reward discounting, but caused interventions by human teammates with low initial trust. In contrast, the robot executing the trust-POMDP policy first removed the three sealed water-bottles to develop trust, and only attempted to remove the glass cup at the end, resulting in fewer interventions and higher overall reward.

%\section{Background}
\section{Related work}
\label{sec:background}

% The important role of trust in the social science literature
%\snnote{I say we skip it all together, and we focus only on trust in automation, since it is a very general topic}
Trust has been studied extensively in the social science research literature~\citep{golembiewski1975centrality, kramer1995trust}, with Mayer et al., suggesting that three general levels summarize the bases of trust: ability, integrity, and benevolence~\citep{mayer1995integrative}. Trust in automation differs from trust between people in that automation lacks intentionality~\citep{lee2004trust}. Additionally, in a human-robot collaboration task, human and robot share the same objective metric of task performance. Therefore, similar to previous work~\citep{desai2012modeling, xu2016towards, pierson2016adaptive, pippin2014trust,wang2016trust}, we assume that human teammates will not expect the robot to deceive them on purpose, and their trust will depend mainly on the \em perceived robot ability \em to complete the task successfully.

Binary measures of trust~\citep{hall1996trusting}, as well as continuous measures~\citep{lee1992trust, desai2012modeling, xu2016towards}, and ordinal scales~\citep{muir1990operators, hoffman2013evaluating} have been proposed. For real-time measurement, ~\citet{desai2012modeling} proposed the Area Under Trust Curve (AUTC) measure, which was recently used to account for one's entire interactive experience with the robot~\citep{yang2017}.  

%Several measurement scales of trust in a robot has been proviously proposed and evaluated, including 
 
Researchers have also studied the temporal dynamics of trust conditioned on the task performance:  \citet{lee1992trust} proposed an autoregressive moving average vector form of time series analysis; \citet{Floyd2015} used case-based reasoning;  \citet{xu2015optimo} proposed an online probabilistic trust inference model to estimate a robot's trustworthiness;  
\citet{wang2016trust} showed that adding transparency in the robot model by generating explanations improved trust and performance in human teams;
\citet{desai2013impact,desai2012effects} showed that robot failures
had a negative impact on human trust, and early robot failures led to 
dramatically lower trust than later robot failures.
While previous works have focused on either quantifying trust or studying the 
dynamics of trust in human-robot interaction, our work enables the robot to leverage upon a model of human trust and choose actions to maximize  task performance. 
% Since we focus on the evolution of trust in a sequential decision making task with distinct task-steps, we chose Muir's questionnaire~\citep{desai2012modeling, muir1990operators} as a validated ground-truth measure of human trust.

%\footnote{Decision-theoretic models of trust have been proposed for software agents (e.g. \citep{norman2011trust}) and used in the context of deciding who and when to trust.}

% planning effective actions for the robot and existing gap
% \subsection{Robot Planning in Human-Robot Collaboration}
In human-robot collaborative tasks, the robot often needs to reason over the human's hidden mental state in its decision-making. The POMDP provides a principled general framework for such reasoning. It has enabled robotic teammates to coordinate through communication~\citep{barrett2014communicating} and software agents to infer the intention of human players in game AI applications~\citep{macindoe2012pomcop}. 
The model has been successfully applied to real-world tasks, such as autonomous driving where the robot car interacts with pedestrians and human drivers~\citep{bai2015intention, bandyopadhyay2013intention,galceran2015multipolicy}. 
When the state and action space of the POMDP model become continuous, one can use hindsight optimization~\citep{javdani2015shared}, or value of information heuristics~\citep{sadigh2016planning}, which generate approximate solutions but are computationally more efficient.

\citet{nikolaidis2015efficient} proposed to infer the human type or preference online using models learned from joint-action demonstrations. This formalism recently extended from one-way adaptation (from robot to human) to human-robot \em mutual \em adaptation~\citep{nikolaidis2016formalizing,nikolaidis2017formalizing}, where the human may choose to change their preference and follow a policy demonstrated by the robot in the recent history. In this work, we provide a general way to link the whole interaction history with the human policy, by incorporating human trust dynamics into the planning framework. 

%While those works have assumed one-way adaptation from robot to human, \citet{nikolaidis2016formalizing} recently proposed a human-robot mutual adaptation formalism, where the human may choose to change their preference and follow a policy demonstrated by the robot in the recent history with some probability equal to the human adaptability. In this work, we provide a general way to link the whole interaction history with the human policy, by incorporating human \em trust \em dynamics into the planning framework. 

%as well as by switching roles with the robot~\citep{nikolaidis2013human}.

% \section{Trust-POMDP for Human-Robot collaboration}
\section{Trust-POMDP}
\label{sec:trust-POMDP}

\subsection{Human-robot team model}
\label{subsec:humanrobotteam}
We formalize the human-robot team as a Markov Decision Process (MDP), with world state $\x{} \in \X$, 
robot action $\ar{} \in \AR$, and human action 
$\ah{} \in \AH$. 
The system evolves according to a probabilistic 
state transition function $p(\x{}'|\x{},\ar{},\ah{})$ which specifies the probability of transitioning from state $\x{}$ to state $\x{}'$ when actions $\ar{} \text{ and }\ah{}$ are applied in state $\x{}$. After transitioning, the team receives a real-valued reward $\rwd{}(\x{},\ar{},\ah{}, \x{}')$, which is constructed to elicit the desirable team behaviors.

We denote by 
\allowbreak $h_t = 
\{ \x{0},    \ar{0}, \ah{0}, 
   \x{1}, \rwd{1}, 
   \hdots, 
   \x{t-1}, \ar{t-1}, \ah{t-1}, \x{t}, \rwd{t}
\} $ $ \in H_t$ as the history of interaction between robot and human until time step $t$.
In this paper, we assume that the human observes the robot's current action  and then decides their own action. 
In the most general setting, the human uses the entire interaction history  $h_t$ to decide the action.
Thus, we can write the human's (possibly stochastic) policy as $\policyh(\ah{t} | \x{t}, \ar{t}, h_t)$ which outputs the probability of each human action $\ah{t}$.

Given a robot policy $\policyr$,  the \emph{value}, \textit{i.e.}, the expected total discounted reward  of starting at a state $\x{0}$ and following the robot and human policies is 
\begin{equation}
v(\x{0}|\policyr, \policyh) =
\underset{\ar{t} \sim \policyr, \ah{t} \sim \policyh} {\mathbb{E}}~
%\underset{\ah{t} \sim \policyh} {\mathbb{E}}
\sum_{t=0}^{\infty}
\gamma^t \rwd{}(\x{t},\ar{t},\ah{t}),
\end{equation}
and the robot's optimal policy \policyropt can be computed as
\begin{equation}
\policyropt = 
\underset{\policyr}{\argmax}~
v(\x{0}|\policyr, \policyh).
\end{equation}

In our case, however, the robot does not know the human policy in advance.  It computes the  optimal policy under expectation over the human policy:
\begin{equation}
\policyropt = 
\underset{\policyr}{\argmax}~
\underset{\policyh}{\mathbb{E}}
v(\x{0}|\policyr, \policyh).
\label{eq:MAMDP}
\end{equation}
Key to solving \equref{eq:MAMDP} is for the robot to model the human policy, which  potentially depends on the entire history $h_t$. The history $h_t$ may grow arbitrary long and make the optimization extremely difficult.

\subsection{Trust-dependent human behaviors}
\label{subsec:humanmodel}

Our  insight is that in a number of human-robot collaboration scenarios,
\emph{trust is a compact approximation of the interaction history~$h_t$}. 
This allows us to condition human
behavior on the inferred trust level and in turn find the optimal policy that maximizes team performance.

Following previous work on trust modeling~\citep{xu2015optimo},
we assume that trust can be represented as a single scaler random variable $\trust{}$. Thus, the human policy is rewritten as 
\begin{equation}
\policyh(\ah{t} | \x{t}, \ar{t}, \trust{t})
=
\policyh(\ah{t} | \x{t}, \ar{t}, h_t).
\end{equation}

% Of course, human trust changes over time, and has its own dynamics.
% Similar to previous work on trust modeling~\citep{xu2015optimo},
% we apply \emph{causal} reasoning to update the robot's deserved
% trustworthiness based on its task performance.
% \begin{align}
% \trust{t+1} \sim p(\trust{t+1}|\trust{t},\x{t+1},\rwd{t+1})
% % \trust{t+1} \sim p(\trust{t+1}|\trust{t},\x{t+1},\perf{t+1})
% \label{eq:trustdyna}
% \end{align}
% where, $\perf{t+1}$ is the performance of the robot at current time step.
% where, $\rwd{t+1}$ is the reward robot get at current time step.
% We detail in \secref{sec:learntrust} how we can learn trust dynamics via interaction.
% \minnote{We decide here whether to use \rwd{t+1} or \perf{t+1}}

\subsection{Trust dynamics}
\label{subsec:trust-performance}

% Our model for trust dynamics in \equref{eq:trustdyna} is very general but to build such a general model requires prohibitive amounts of human data. 
Human trust changes over time. 
% , and possesses its own dynamics. 
We adopt a common assumption on the trust dynamics: trust evolves based on the robot's {performance} $\perf{t}$~\cite{lee1992trust,xu2015optimo}. Performance can depend not just on the current and transitioned world state but also the human and robot's actions 
\begin{equation}
\perf{t+1} = \texttt{performance}(\x{t+1}, \x{t}, \art, \aht).
\end{equation}
For example, \texttt{performance} may indicate success or failure of the robot to accomplish a task. This allows us to write our trust dynamics equation as
\begin{align}
\trust{t+1} \sim p(\trust{t+1}|\trust{t},\perf{t+1})
\label{eq:trustdynb}.
\end{align}
We detail in \secref{sec:learntrust} how trust dynamics is learned via interaction.
 
\subsection{Maximizing team performance}
\label{subsec:trustplanning}

Trust cannot be directly observed by the robot and therefore must be inferred from the human's actions. In addition, armed with a model, the robot may actively modulate the human's trust for the team's long-term reward.

\begin{figure}[t]
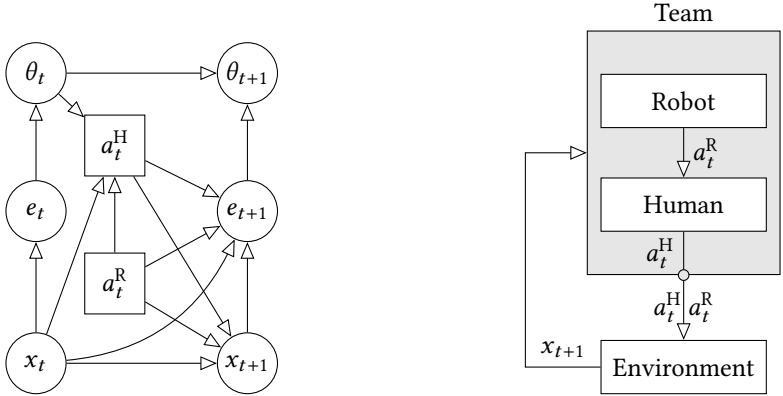

    \centering
    \includestandalone{figs/trust_model}    
    \hspace{3cm}
    \includestandalone{figs/trust_flowchart}    
    \caption{
        The trust-POMDP graphical model (left) and 
        the team interaction flowchart (right). 
        The robot's action $a^R_t$ depends
        on the world state $\x{t}$ and its belief over trust
        $\trust{t}$.  
        %The trust-POMDP graphical model (left) and interaction flowchart (right).
        }
    \label{fig:trustplanning}
\end{figure}

We  achieve this behavior by modeling the interaction as a partially observable Markov decision process (POMDP), which provides a principled  general framework for sequential decision making under uncertainty.
A graphical model of the Trust-POMDP and a flowchart of the interaction are shown in \figref{fig:trustplanning}. 

To build trust-POMDP,
we  create an augmented state space with the augmented state $s = (\x{}, \trust{})$ composed of the fully-observed world state $\x{}$ and the \emph{partially-observed} human trust $\trust{}$. We maintain a belief $b$ over the human's trust. The trust dynamics and human behavioral policy are embedded in the transition dynamics of trust-POMDP. We describe in \secref{sec:learntrust} how we learn the trust dynamics and the human behavioral policy.

The robot now has two distinct objectives through its actions: 
\begin{itemize}
\item \emph{Exploitation.}  Maximize the team's reward 
\item \emph{Exploration.} Reveal and change the human's trust so that future actions are rewarded better.
\end{itemize}

The solution to a Trust-POMDP is a policy that maps belief states to robot actions,
\ie, $\ar{} = \policyr(b_t, x_t)$.
% How do we solve trust-POMDP in this paper
To compute the optimal policy, we use the SARSOP algorithm~\citep{kurniawati2008sarsop}, which is computationally efficient and has been previously used in various robotic tasks~\citep{bandyopadhyay2013intention}.

\section{Learning Trust Dynamics and Human Behavioral Policies}
\label{sec:learntrust}

Nested within the trust-POMDP is a model of human trust dynamics
\allowbreak $p(\trust{t+1}|\trust{t}, \perf{t+1})$, 
and behavioral policy
$\policyh(\ah{t} | \x{t}, \ar{t}, \trust{t})$. We adopted a data-driven approach and built the two models for the table clearing task from data collected in an online AMT experiment. Suitable probabilistic models derived via alternative approaches can be substituted for these learned models (\eg, for other tasks and domains).

%In this section, we demonstrate learning the trust dynamics and human behavioral policy for the table clearing task. % in~\figref{fig:front}. 

\subsection{Data Collection} \label{subsec:DataCollection}

\noindent\textbf{Table clearing task.} 
A human and a robot collaborate to clear objects off a table. The objects include three water bottles, one fish can, and one wine glass. At each time step, the robot picks up one of the remaining objects. Once the robot starts moving towards the intended object, the human can choose between two actions: \{intervene and pick up the object that the robot is moving towards, stay put and let the robot pick the object by itself\}. This process is repeated until all the objects are cleared from the table. 

% In this scenario, we model the trust dynamics to depend only on the performance of the robot $p(\trust{t+1}|\trust{t},\x{t+1},\rwd{t+1}) \equiv p(\trust{t+1}|\trust{t},\perf{t+1})$. 
Each object is associated with a different reward, based on whether the robot successfully
clears it from the table (which we call SP-success), the robot fails in clearing it 
(SP-fail), or the human intervenes and puts it on the tray (IT). 
Table~\ref{table:table-clearing-payoff} shows the rewards for each object and outcome. 
We assume that a robot success is always better than a human intervention, 
since it reduces human effort. Additionally, there is no penalty if the robot fails 
by dropping one of the sealed water bottles, since the human can pick it up. 
On the other hand, dropping the fish can result in some penalty, since its contents 
will be spilled on the floor. Breaking the glass results in the highest penalty. 
We see that staying put when the robot attempts to pick up the bottle has the lowest risk, 
since there is no penalty if the robot fails. On the other hand, staying put in the case 
of the glass object has the largest risk-return trade off.  We expect the human to let 
the robot pick up the bottle even if their trust is low, since there is no penalty if the 
robot fails. On the other hand, if the human does not trust the robot, we expect them to 
likely intervene on glass or can, rather than risking a high penalty in case of robot failure.
%When we conduct the experiment, we assume that the robot never fails in the 
%table clearing task, 
%and this information is unknown to the participants. 
%One can use the same approach as described to learn trust dynamics 
%for robot failures as well.

In this work, we choose the table clearing task to test our trust-POMDP model, 
because it is simple and allows us to analyze
experimentally the core technical issues on human trust without interference
from confounding factors.
Note that the primary objective and contribution of this work are to develop
a mathematical model of trust embedded in a decision framework, and to show
that this model improves human robot collaboration.
In addition, we believe that the overall technical approach in our work
is general and not restricted to this particular simplified task.
What we learned here on the trust-POMDP for a simplified task will be
a stepstone towards more complex, large-scale applications.

\noindent\textbf{Participants.} For the data collection, we recruited in total $231$ participants through Amazon's Mechanical Turk (AMT)~\footnote{We conducted two sessions of data collection, one where the robot always succeeded and one when the robot failed with high probability. Our previous work~\cite{chen2018planning} presents the results of the first session only.}. The participants are all from United States, aged 18-65 and with approval rate higher than $95\%$. Each participant was compensated $\$1$ for completing the study. To ensure the quality of the recorded data, we asked all participants an attention check question that tested their attention to the task. We removed $9$ data points either because the participants failed on the attention check question or the their data were incomplete. This left us $222$ valid data points for model learning.

% the table shows the payoff of the task
\newcommand{\specialcell}[2][c]{%
  \begin{tabular}[#1]{@{}c@{}}#2\end{tabular}}

\newcolumntype{C}[1]{>{\centering\let\newline\\\arraybackslash\hspace{0pt}}m{#1}}
\newcolumntype{R}{>{\raggedleft\arraybackslash}p{2cm}}
\newcolumntype{L}{>{\raggedright\arraybackslash}p{1cm}}
\begin{table}
\caption{The reward function $R$ for the table-clearing task.}
\label{table:table-clearing-payoff}
%\vspace*{-6pt}
 %\begin{tabular}{C{1cm}|C{2cm}|C{2cm}|C{2cm}|}
 \begin{tabular}{L R R R }
 \hline
 
 \hline
     &Bottle & Fish Can & Wine Glass    %\vspace{0.5em}
   \\ \hline
  \specialcell{SP-success} & $1$ & $2$ & $3$ \\ 
  \specialcell{SP-fail} & $0$ & $-4$ & $-9$ \\ 
  \specialcell{IT} & $0$ & $0$ & $0$ \\ \hline
 \end{tabular}
 %\vspace{-0.2cm}
\end{table}

\noindent\textbf{Procedure.} Each participant is asked to perform an online table clearing task together with a robot. 
Before the task starts, 
the participant is informed of the reward function in~\tabref{table:table-clearing-payoff}.
We first collect the participant's initial trust in the robot.
We used Muir's questionnaire~\citep{muir1990operators}, with a seven-point Likert scale as a human trust metric, \ie, trust ranges from $1$ to $7$. The Muir's questionnaire we used is listed in~\tabref{table:muir}. At each time step, the participant watches a video of the robot attempting to pick up an object, and are asked to choose to intervene or stay put. They then watch a video of either the robot picking up the object, or them intervening based on their action selection. Then, they report their updated trust in the robot. 

\begin{table}
\caption{Muir's questionnaire.}
\label{table:muir}
%\vspace*{-6pt}
\begin{tabular}{ l }
\hline

\hline
1. To what extent can the robot's behavior be predicted from\\
moment to moment? \\
2. To what extent can you count on the robot to do its job? \\
3. What degree of faith do you have that the robot will be able\\
to cope with similar situations in the future? \\
4. Overall how much do you trust the robot? \\
\hline
\end{tabular}
%\vspace{-0.2cm}
\end{table}

We are interested in learning the trust dynamics and the human behavioral 
policies for any state and robot action. However, the number of
open-loop~\footnote{When collecting data from AMT, the robot follows an open-loop
    policy, \ie, it does not adapt to the human behavior.}
robot policies is $O(K!)$, where $K$ is the number of objects on the table.
In order to focus the learning on a few interesting robot policies (i.e. picking up the glass in the beginning vs in the end), while still covering a large space of policies, we split the data collection process, so that in one half of the trials the robot randomly chooses a  policy out of a set of pre-specified policies, while in the other half the robot follows a random policy.

\noindent\textbf{Data Format.} The data we collected from each participant has the following format:
\begin{align*}
    d_i = \{ \trust{0}^\textrmsmall{M}, a^{\textrmsmall{R}}_0, a^{\textrmsmall{H}}_0, \perf{1}, 
    \trust{1}^{\textrmsmall{M}}, \hdots, 
    a^{\textrmsmall{R}}_{K-1}, a^{\textrmsmall{H}}_{K-1}, \perf{K}, 
    \trust{K}^{\textrmsmall{M}}\}
\end{align*}
where $K$ is the number of objects on the table. 
$\trust{t}^{\textrmsmall{M}}$ is the estimated human trust at time $t$ 
by averaging the participants' responses to the 
Muir's questionnaire to a single rating between 1 and 7. $a^{\textrmsmall{R}}_t$ is the 
action taken by the robot at time step $t$. $a^{\textrmsmall{H}}_t$ is the action taken 
by the human at time step $t$. $\perf{t+1}$ is the performance of the robot that indicates 
whether the robot succeeded at picking up the object, the robot failed, or the human intervened.

%takes the following values in our table
%e \{human intervenes on the water bottle, human stay put on the water bottle and robot
%succeed, human intervenes on the fish can, human stay put on the fish can and robot succeed, 
%human intervenes on the glass cup, human stay put on the glass cup and robot succeed\}.
%$\trust_{t+1}$ is the updated human trust after saw the robot's performance $f_{t+1}$.

\subsection{Trust dynamics model} \label{subsec:trust-dynamics}

We model human trust evolution as a linear Gaussian system.
Our trust dynamics model relates the human trust causally to the robot task performance 
$\perf{t+1}$.
%\eg, robot successfully picks up the glass cup or human intervenes.

\begin{align}
    & P(\trust{t+1} | \trust{t}, \perf{t+1}) = \mathcal{N} (\alpha_{\perf{t+1}} \trust{t} + \beta_{\perf{t+1}}, \sigma_{\perf{t+1}}) \\
    & \trust{t}^M \sim \mathcal{N} (\trust{t}, \sigma^2) \ , \ \trust{t+1}^M \sim \mathcal{N} (\trust{t+1}, \sigma^2)
\end{align}
where $\mathcal{N}(\mu, \sigma)$ denotes a Gaussian distribution with mean
$\mu$ and standard deviation $\sigma$.
$\alpha_{\perf{t+1}}$ and $\beta_{\perf{t+1}}$ are linear coefficients for the trust
dynamics, given the
robot task performance $\perf{t+1}$.
In the table clearing task, $\perf{t+1}$ indicates whether the robot
succeeded at picking up an object, the robot failed, or the human intervened,  \eg, $\perf{t+1}$ can represent that 
the robot succeeded at picking a water bottle, or that the human intervened at
the wine glass. 
$\trust{t}^M$ and $\trust{t+1}^M$ are the observed human trust (Muir's questionnaire) 
at time step $t$ and time step $t+1$. 

\begin{figure}%[t!]
    \centering
    \resizebox{\columnwidth}{!}
    {\includegraphics{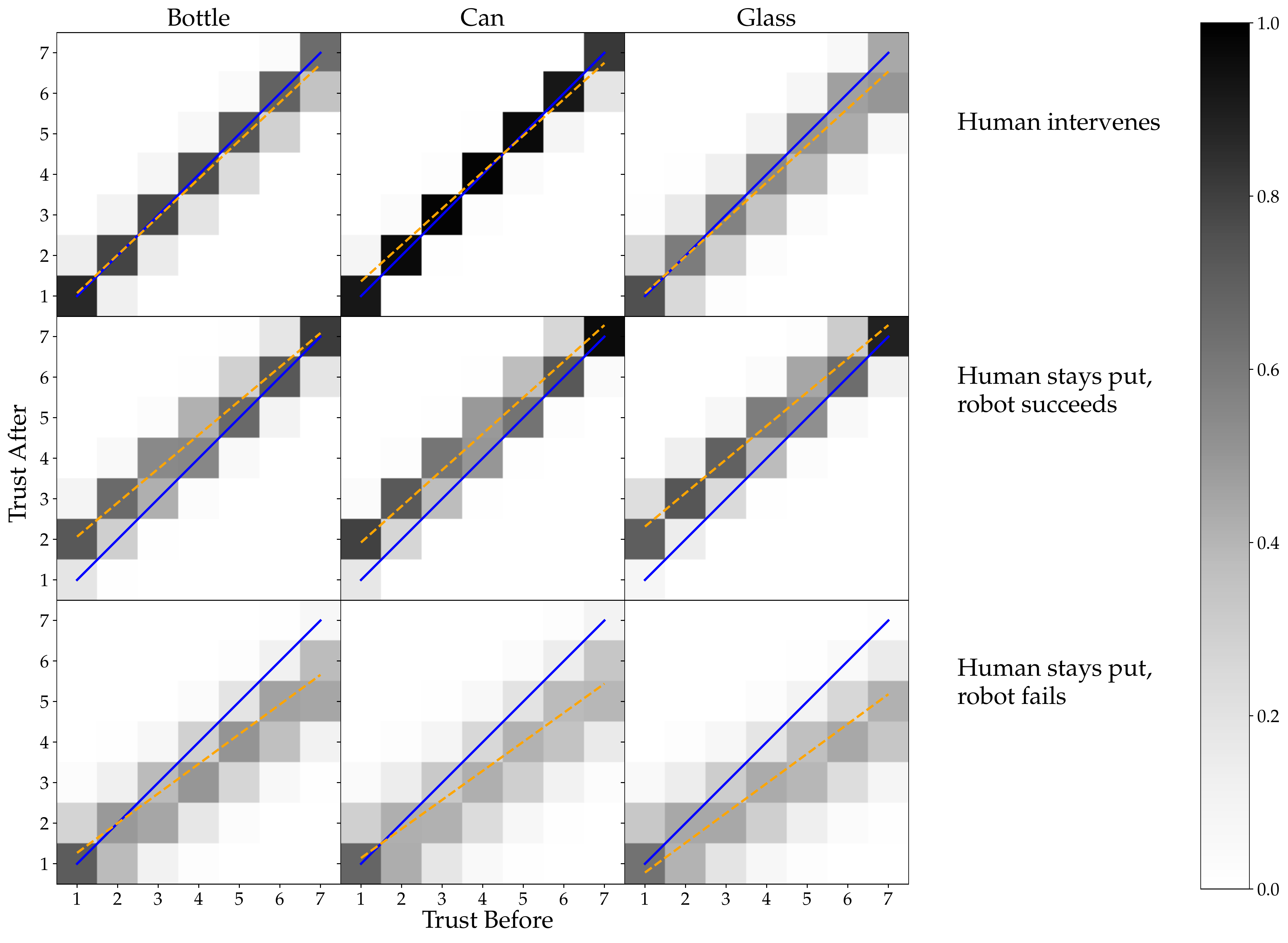}}
    \caption{
        %Top / center row: 
        Trust transition matrices, which represent the change of trust given the robot performance, shown by the linearly regressed line (yellow) contrasted with the X-Y line (blue).
%         The $y$-axis represents the trust value before the robot took its action, 
%         and $x$ axis represents the trust value after the human saw the robot's performance. 
%         The title represents the performance of the robot, \eg, ``Bottle, Intervene" represents 
%         the case where human intervenes on the water bottle. 
In general, trust stays constant or decreases slightly when the human intervenes (top row). It increases when the human stays put and the robot succeeds (middle row), while it decreases when the robot fails (bottom row). }
    \label{fig:trustdyna}
    %\vspace{-0.4cm}
\end{figure}

% model fitting
The unknown parameters in the trust dynamics model include $\alpha_{\perf{t+1}}$, $\beta_{\perf{t+1}}$, $\sigma_{\perf{t+1}}$ and $\sigma$.
We performed full Bayesian inference on the model through Hamiltonian Monte Carlo
sampling using the Stan probabilistic programming platform~\cite{carpenter2016stan}.
\figref{fig:trustdyna} shows the trust transition matrices for all possible robot performance
in the table clearing task.
As we can see, human trust in the robot gradually increased with observations of successful robot actions (as indicated by transitions to higher trust levels when the participants stayed put and robot succeeded), while it decreased with observations of robot failures. Trust tended to remain constant or decrease slightly when interventions occurred. It also appears that that the higher the trust, the greater the loss upon failure, and vice versa upon success. These results matched our expectations that successful robot performance positively influenced trust, while robot failures negatively affected trust. 

\subsection{Human behavioral policies} 
\label{subsec:human-behavioral-prediction}

% clarify why we need those two human models
Our key intuition in the human model is that human's behavior depends on 
the trust in the robot. To support our intuition, we consider two types of human behavioral models.
The first model is a trust-free human behavioral model that ignores human trust,
while the second is a trust-based human behavioral model that explicitly models 
human trust.
In both human models, we assume humans
follow the \em softmax rule \em~\footnote{According to the \em softmax rule \em, 
    the human's decision
    of which action to take is determined probabilistically on the 
    actions' relative expected values.}
when they make decisions in an uncertain
environment~\citep{daw2006cortical}.
More explicitly,

\begin{itemize}
    \item Trust-free human behavioral model: 
        At each time step, the human 
        %follows the \em softmax rule \em
        %by selecting 
        selects an action probabilistically
        based on the actions' relative expected values.
        The expected value of an action depends on
        the human's belief on the robot to succeed
        and the risk of letting robot to do the task. 
        In the trust-free human model,
        the human's belief on the robot success on a particular task
        does not change over time.

    \item Trust-based human behavioral model: 
        Similar to the model above, 
        the human follows the \em softmax rule \em 
        at each time step.
        However, the trust-based human model assumes that 
        human's belief on the robot success 
        changes over time, and it depends on human's trust
        in the robot.
\end{itemize}

%We formally introduce the two human behavioral models as follows.
Before we introduce the models,
we start with some notations.
Let $j$ denote the object that the robot tries to pick at time step $t$.
Let $r_j^{\textrmsmall{S}}$ be the reward if the human stays put and the robot succeeds,
and $r_j^{\textrmsmall{F}}$ be the reward if the human stays put and the robot fails.
Let $\trust{t}$ be the human trust in the robot at time step $t$.
$\mathcal{S}(x) = \frac{1}{1 + \mathrm{e}^{-x}}$ is the sigmoid function,
which is equivalent to the softmax function
in the case of binary human actions.
$\mathcal{B}(p)$ is the Bernoulli distribution that takes 
action stay put
with probability $p$.

The trust-free human behavioral model is as follows,
\begin{align}
    & P_t = \mathcal{S}(b_j r_j^{\textrmsmall{S}} + 
    (1 - b_j) r_j^{\textrmsmall{F}}) \\
    & a_t^{\textrmsmall{H}} \sim \mathcal{B}(P_t)
\end{align}
where, $b_j$ is the human's belief on the robot successfully picking up object $j$,
and it remains constant.
$0 < P_t < 1$ is the probability that human stays put at time step $t$.
$a_t^{\textrmsmall{H}}$ is the action human taken at time step $t$.

Next, we introduce the trust-based human behavioral model:

\begin{align}
    & b^t_j = \mathcal{S}(\gamma_j \trust{t} + \eta_j) \\
    & P_t = \mathcal{S}(b^t_j r_j^{\textrmsmall{S}} + 
    (1 - b^t_j) r_j^{\textrmsmall{F}}) \\
    & \trust{t}^{M} \sim \mathcal{N}(\trust{t}, \sigma^2) \ , \ a_t^{\textrmsmall{H}} \sim \mathcal{B}(P_t)
\end{align}

where $b^t_j$ is the human's belief on robot success on object $j$ at time step $t$,
and it depends on the human's trust in the robot. 
$\gamma_j$ and $\eta_j$ are the linear coefficients for object $j$.
$0 < P_t < 1$ is the probability that the human stays put at time step $t$.
$\trust{t}^{M}$ is the observed human trust from Muir's questionnaire at time step $t$, 
and we assume it follow a Gaussian distribution with mean $\trust{t}$
and standard deviation $\sigma$. 
$a_t^{\textrmsmall{H}}$ is the action human taken at time step $t$.

% log-likelihood of when fitting the two models above, and the one-way anova score
The unknown parameters here include 
$b_j$ in the trust-free human model, and $\gamma_j$, $\eta_j$, 
$\sigma$ in the trust-based human model.
We performed Bayesian inference on the two models above using
Hamiltonian Monte Carlo sampling~\cite{carpenter2016stan}.
The trust-based human model (log-likelihood $= -153.37$) fit the collected AMT data 
better than the trust-free human model (log-likelihood $= -156.40$). The log-likelihood values are relatively low in both two models 
due to the large variance among different users. Nevertheless, this result supports our notion that the prediction on human behavior is 
improved when we explicitly model human trust.

\begin{figure}[t!]
    \centering
    \resizebox{0.8\columnwidth}{!}
    {\includegraphics{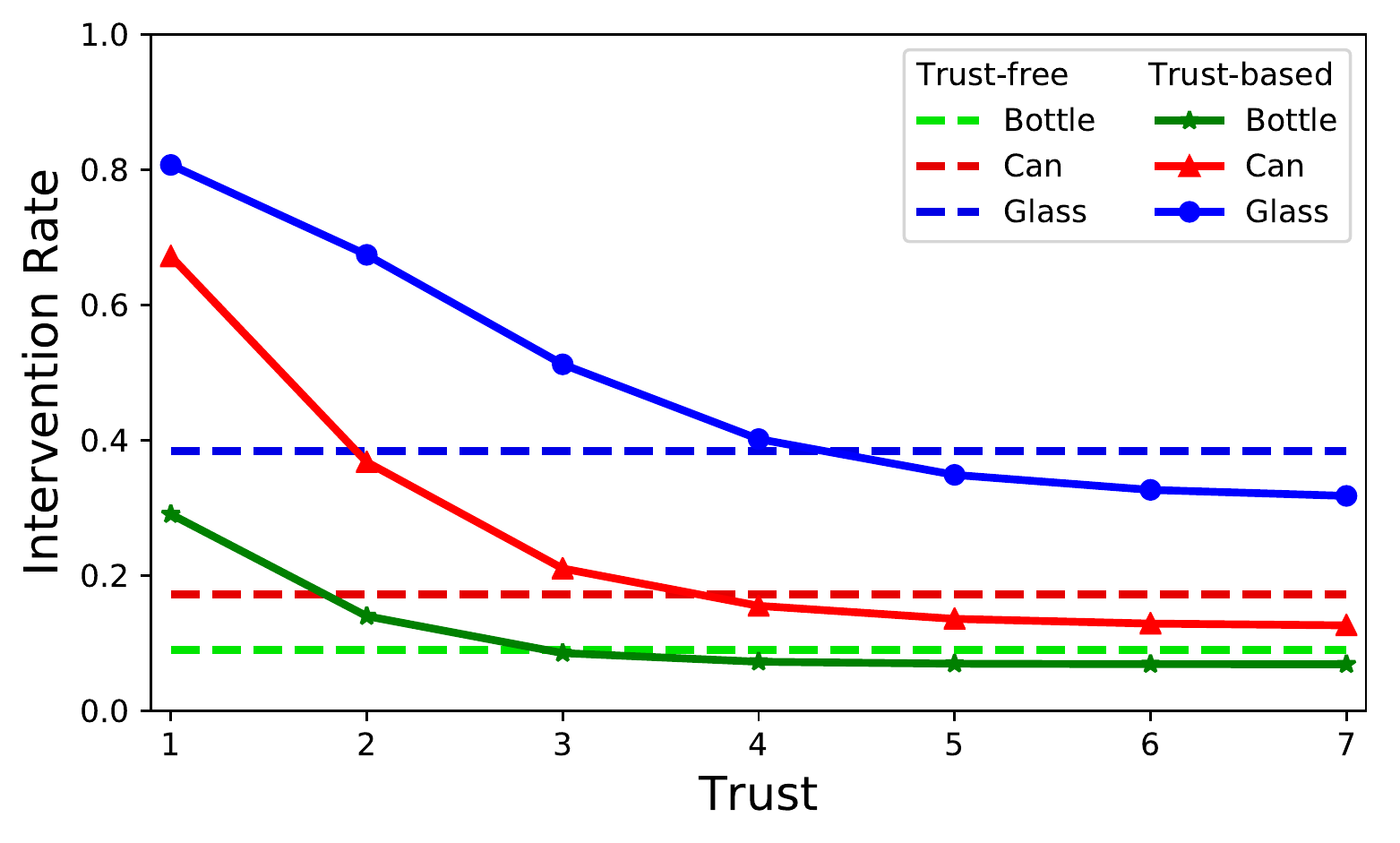}}
    \caption{
        %The change of human intervention rate with respect to trust.
        The model prediction on the mean of human intervention rate with respect
        to trust.
        Under the trust-free human behavioral model,
        which does not account for trust,
        the human intervention rate stays constant. 
        Under  the trust-based human behavioral model, 
        the intervention rate decreases with increasing trust. 
        The rate of decrease depends on the object; it is more sensitive to the
        risker objects.
    }
    \label{fig:humanpoly}
    %\vspace{-0.4cm}
\end{figure}

% Plot the human behavioral policy for the two models above
\figref{fig:humanpoly} shows the mean probability of human interventions 
with respect to human's trust in the robot.
For both models, the human tends to intervene more on objects with higher risk,
\ie, the human intervention rate on glass is higher than can, which is again higher than bottle.
% the human intervention rate on can is higher than bottle.
The trust-free human behavioral model ignores human trust, 
thus the human intervention rate does not change.
On the other hand, the trust-based human behavioral model has a general falling trend,
which indicates that participants are less likely to intervene when their trust
in the robot is high.
This is observed particularly for the highest-risk object (glass), where the object 
intervention rate drops significantly when human trust score is maximum.  

To summarize, the results of Sec.~\ref{subsec:trust-dynamics} and 
\secref{subsec:human-behavioral-prediction} indicate that 
\begin{itemize}
    \item Human trust is affected by robot performance:
        human trust can be built up by successfully picking up objects (\figref{fig:trustdyna}).
        In addition, it is a good strategy for the robot to start with low risk 
        objects (bottle), since the human is less likely to intervene even if 
        the trust in the robot is low (\figref{fig:humanpoly}).
    \item Human trust affects human behaviors: 
        the intervention rate on the high risk objects could
        be reduced by building up human trust (\figref{fig:humanpoly}).
\end{itemize}

%\begin{figure*}[t!]
%\vspace{-0.5cm}
 %\centering
  %\includegraphics[width=1.0\linewidth]{figs/performance_figs.pdf}
 %\caption{Left: Intervention rate for each policy and object type.  Center: Mean trust score annotated by participants over time. Right: Intervention rate per annotated trust score for all participants. All error bars denote standard error of the mean.}
  %\label{fig:performance_figs}
 %\end{figure*}

\section{Experiments}
\label{sec:experiment}
We conducted two human subjects experiments, one on AMT with human participants interacting
with recorded videos and one in our lab with human participants interacting with a real robot.
%\footnote{We refer the reader to~\citet{bethel2010review} for a review of methods for hypothesis testing in  human-robot interaction.} 
The purpose of our study was to test whether the trust-POMDP robot policy would result 
in better team performance than a policy that did not account for human trust. To simplify the analysis of the different behaviors in these experiments, we had the robot always succeed when attempting to pick up the objects.  

%\subsection{Independent Variables}
We had two experimental conditions, which we refer to as ``trust-POMDP'' and ``\baseline''. 

\begin{itemize}
    \item In the trust-POMDP condition, the robot uses human trust as a means to optimize the long term team performance. It follows the policy computed from the trust-POMDP described in ~\secref{subsec:trustplanning}, where the robot's perceived human policy is modeled via the trust-based human behavioral model described in~\secref{subsec:human-behavioral-prediction}. 
%     \snnote{changed it to make clear that the difference is in the model instead of the actual human policies followed}
    \item In the \baseline condition, 
    the robot ignores human trust. It follows a myopic policy by
    optimizing \equref{eq:MAMDP}, where the robot's perceived
    human policy is modeled via the trust-free human behavioral model
    described in~\secref{subsec:human-behavioral-prediction}.
\end{itemize}

\subsection{Online AMT experiment}

%\vspace{0.5em}
\noindent\textbf{Hypothesis 1.}~\textit{ In the online experiment, the performance of teams in the trust-POMDP condition will be better than of the teams in the \baseline condition.} 

\noindent We evaluated team performance by the accumulated reward over the task. We expected the trust-POMDP robot to reason over the probability of human interventions, and act so as to minimize the intervention rate for the highest reward objects.
The robot would do so by actively building up human trust before it goes for high risk objects.
On the contrary, the \baseline robot policy was agnostic to how the human policy may change from the robot and human actions.

%\vspace{0.5em}
\noindent\textbf{Procedure.}
The procedure is similar to the one for data collection (Sec.~\ref{subsec:DataCollection}), with the difference that, rather than executing random sequences, the robot executes the policy associated with each condition. While we kept the Muir's questionnaire in the experiment as a groundtruth measure of trust, the robot \em did not use the score\em, but estimated trust solely from the trust dynamics model as described in Sec.~\ref{subsec:trust-dynamics}.

%\vspace{0.5em}
\noindent\textbf{Model parameters.} In the formulation of \secref{subsec:trustplanning}, the observable state variable $x$ represents the state of each object (on the table or removed). We assume a discrete set of values of trust $\trust{}$ : $\{1,2,3,4,5,6,7\}$. The transition function incorporates the learned trust dynamics and human behavioral policies, as described in Sec.~\ref{sec:learntrust}. The reward function $R$ is given by Table~\ref{table:table-clearing-payoff}. We used a discount factor of $\gamma = 0.99$, 
which favors immediate rewards over future rewards.
%reflecting that with everything else being equal, objects with high reward (glass) should be removed first.
%The total size of the POMDP state-space was $7 \times 32 = 224$ states.

%\vspace{0.5em}
\noindent\textbf{Subject Allocation} We chose a between-subjects design in order to not bias the users with policies from previous conditions. We recruited 208 participants through Amazon Mechanical Turk, aged $18-65$ and with approval rate higher than $95\%$.
Each participant was compensated $\$1$ for completing the study. 
We removed $7$ wrong (participants failed on the attention check question) or incomplete data points. In the end, we had $101$ data points for the trust-POMDP condition, and $100$ data points for the \baseline condition.

\subsection{Real-robot experiment}
In the real-robot experiment we followed the same robot policies, model parameters and procedures as the online AMT experiment, with that the participants interacted with an actual robot in person. 
%\vspace{0.5em}

\noindent\textbf{Hypothesis 2.}~\textit{ In the real-robot experiment, the performance of teams in the trust-POMDP condition will be better than of the teams in the \baseline condition.} 

%\vspace{0.5em}
\noindent\textbf{Subject Allocation.} We recruited 20 participants from our university, aged 21-65. Each participant was compensated $\$10$ for completing the study. 
% \snnote{if you add this, it may be good to add compensation information for all studies.} 
All data points were kept for analysis, \ie, $10$ data points for the trust-POMDP condition and $10$ data points for the \baseline condition.

%\subsection{Analysis} 
%\label{sec:ResultsAndDiscussion}
\subsection{Team performance}
\label{subsec:performance}
%\noindent\textbf{One-way ANOVA test}
%We evaluated team performance by the accumulated reward of the task. 
We performed an one-way ANOVA test of the accumulated rewards (team performance). In the online AMT experiment, the accumulated rewards of trust-based condition was significantly larger than the \baseline condition \allowbreak $(F(1,199) = 7.81, p = 0.006)$. This result supports Hypothesis 1. 

Similarly, the accumulated rewards of the trust-based condition was significantly larger than the \baseline condition 
\allowbreak $(F(1,18) = 11.22,$ $ p = 0.004)$. This result supports Hypothesis 2. 

%A one-way ANOVA showed that the accumulated reward of teams in the trust-based 
%condition was significantly larger than for teams in the baseline condition 
%($F(1,199) = 7.81, p = 0.006)$, in support of our hypothesis. 

\begin{figure*}[t!]
     \centering
     \begin{subfigure}[b]{1.0\linewidth}
         \centering
         \textbf{Online AMT experiment}\par\medskip
         \includegraphics[height=0.20\linewidth]{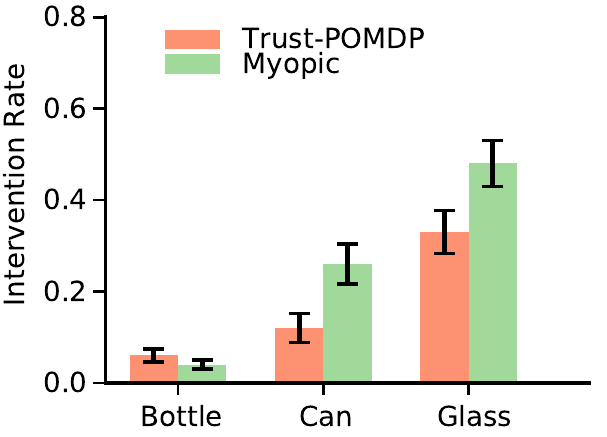}
         \includegraphics[height=0.20\linewidth]{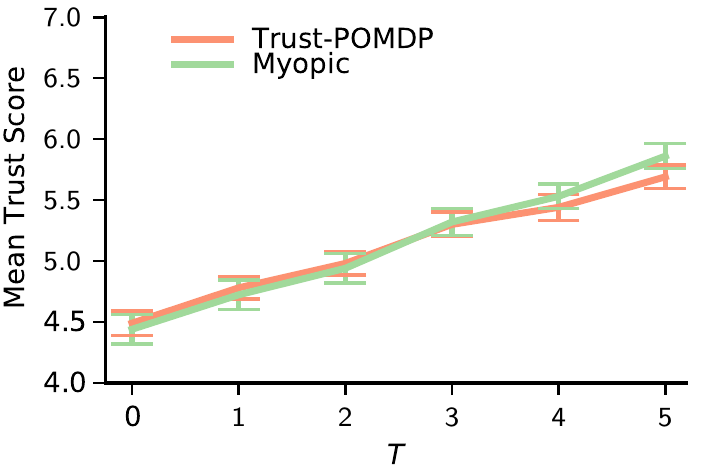}
         \includegraphics[height=0.20\linewidth]{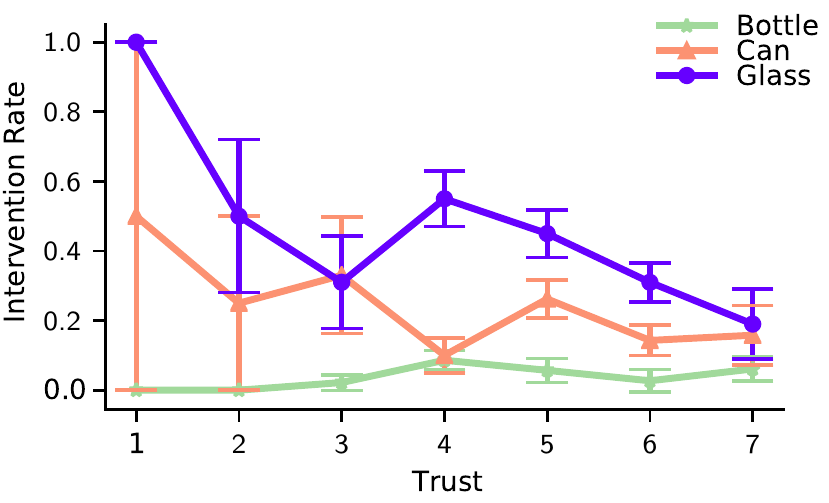}
%          \caption{Amazon mechanical turk (AMT) study}
%          \label{fig:AMTstudy}
    \end{subfigure}
    ~\\
    ~\\
    \begin{subfigure}[b]{1.0\linewidth}
        \centering   
        \textbf{Real-robot experiment}\par\medskip
        \includegraphics[height=0.20\linewidth]{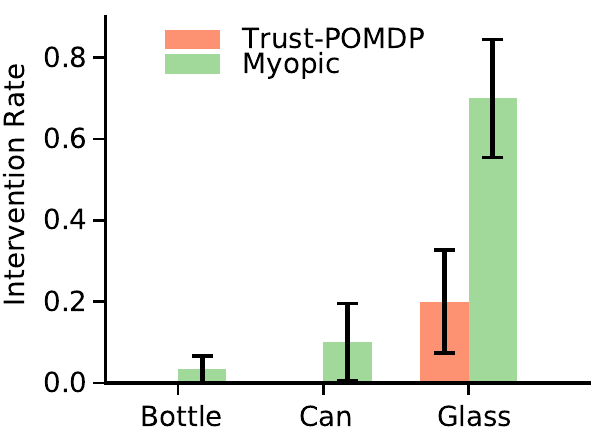}
        \includegraphics[height=0.20\linewidth]{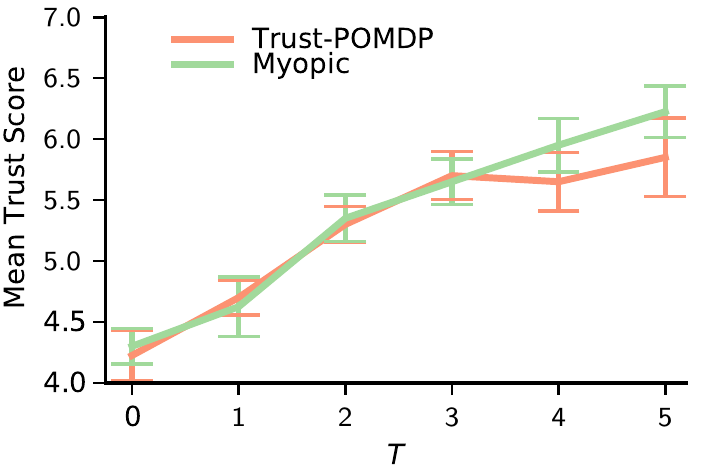}
        \includegraphics[height=0.20\linewidth]{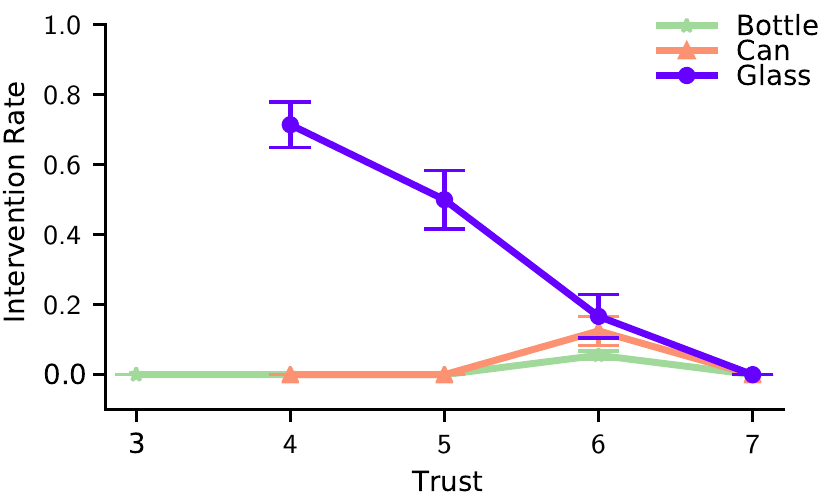}
%         \caption{Lab study}
%         \label{fig:Labstudy}
    \end{subfigure}
    \caption{Comparison of the Trust-POMDP and the myopic policies in the AMT experiment and the real-robot experiment.
%     Left: Intervention rate for each policy and object type.  Center: Mean trust score annotated by participants over time. Right: Intervention rate per annotated trust score for all participants. All error bars denote standard error of the mean.
    }
    \label{fig:performance_figs}
    %\vspace{-0.2cm}
 \end{figure*}

The difference in performance occurred because participants' intervention rate in 
the trust-POMDP condition was significantly lower than \baseline 
condition (\figref{fig:performance_figs} - left column).
In the online AMT experiment, the intervention rate in the trust-POMDP condition was
54\% and 31\% lower in the can and glass object.
In the real-robot experiment, the intervention rate in the trust-POMDP condition dropped to zero (100\% lower) in the can object and 71\% lower in the glass object.

In the \baseline condition, the robot picked the objects 
in the order of highest to lowest reward (Glass, Can, Bottle, Bottle, Bottle). In contrast, the trust-based human behavior model influenced 
the trust-POMDP robot policy by capturing the fact that interventions on high-risk objects were more likely if trust in the robot was insufficient. 
Therefore, the trust-POMDP robot reasoned that it was better to start with the
low risk objects (bottles), 
build human trust (\figref{fig:performance_figs} - center column) and go for high risk object
(glass) last.
In this way, the trust-POMDP robot minimized the human intervention ratio on
the glass and can object, which significantly improved the team performance.

%since the penalty due to discounting was negligible relative to the gains obtained by preventing unnecessary interventions. 

 \begin{figure}[t!]
     \centering
     \begin{subfigure}[b]{1.0\linewidth}
         \centering
         \fontsize{7}{5}\selectfont
         \textbf{Online AMT experiment}\par\medskip
         \includegraphics[width=0.34\linewidth]{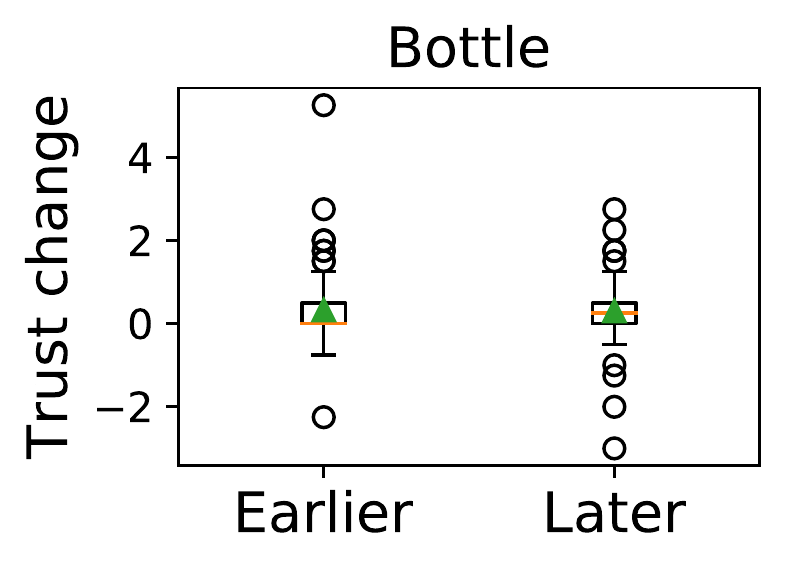}
         \includegraphics[width=0.315\linewidth]{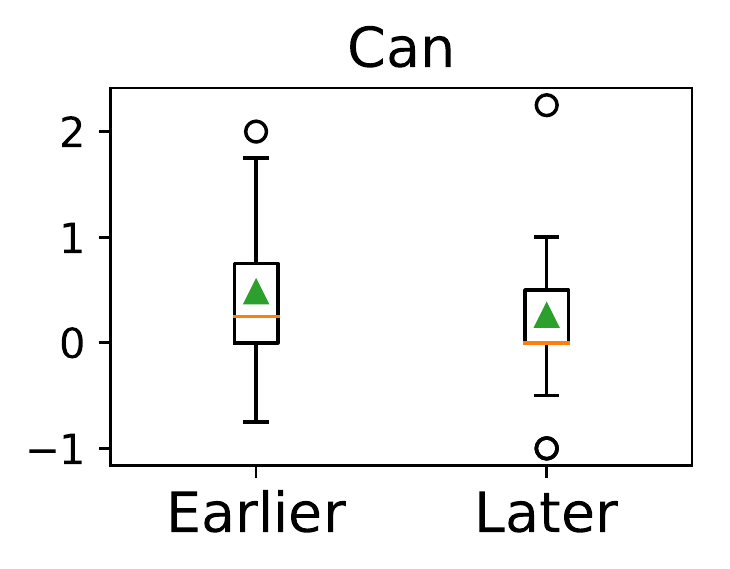}
         \includegraphics[width=0.30\linewidth]{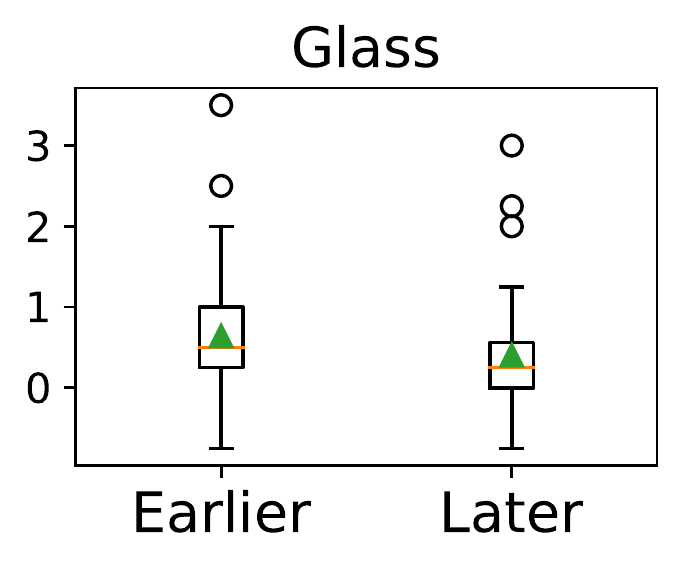}
%          \caption{AMT study}
%          \label{fig:AMTstudy}
    \end{subfigure}
    ~\\
    ~\\
    \begin{subfigure}[b]{1.0\linewidth}
        \centering
        \fontsize{7}{5}\selectfont
        \textbf{Real-robot experiment}\par\medskip
        \includegraphics[width=0.33\linewidth]{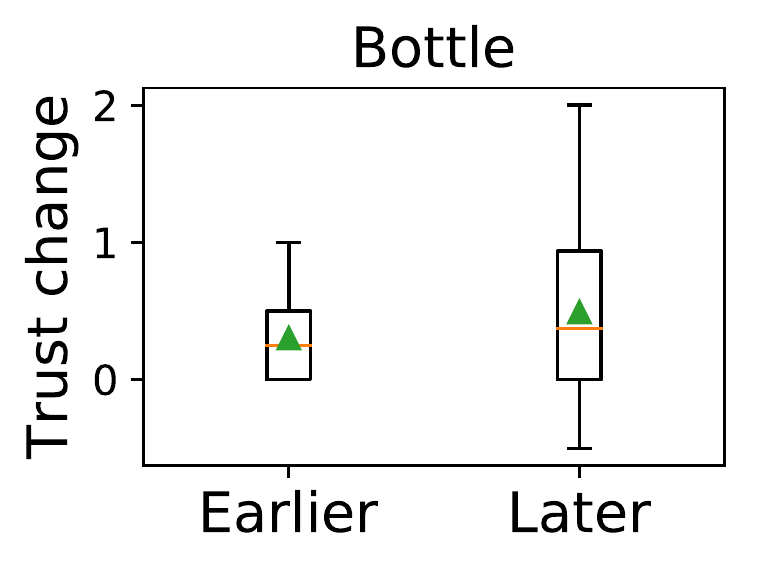}
        \includegraphics[width=0.32\linewidth]{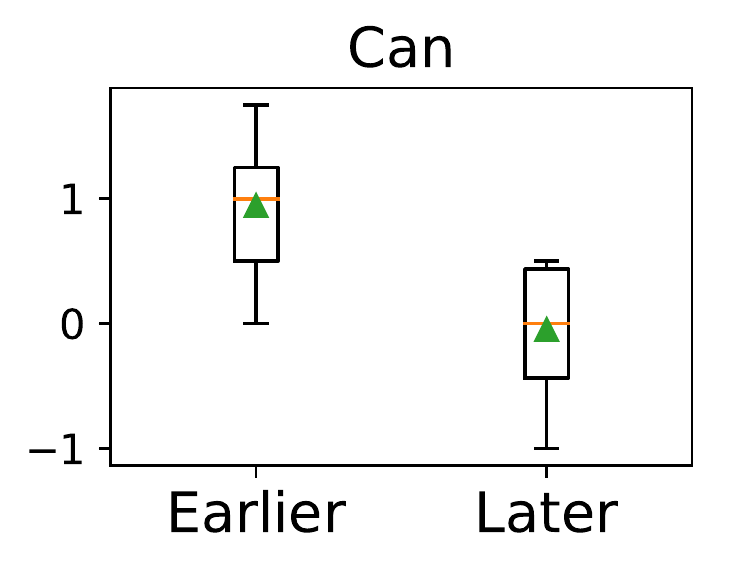}
        \includegraphics[width=0.32\linewidth]{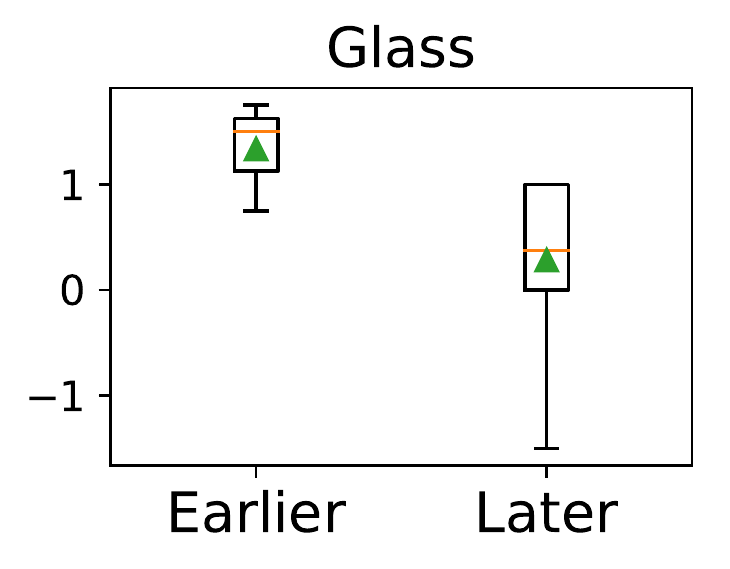}
%         \caption{Lab study}
%         \label{fig:Labstudy}
    \end{subfigure}
    \caption{Time-dependent nonlinear effects of trust dynamics.
        The same outcome has greater effect on trust when it occurs 
        earlier than later.
}
    \label{fig:timeEffectonTrust}
    %\vspace{-0.4cm}
 \end{figure}

%\subsection{Trust evolution and how it affects human behaviors}
\subsection{Trust evolution}
\label{subsec:TrustEvolution}

% Take fig6-center as granted, what does it imply on trust-POMDP
\figref{fig:performance_figs} (center column) shows the participants' trust evolution.
We make two key observations.
First, successfully completing a task increased participants' trust in the robot.
This is consistent with the human trust dynamics model we learned
in~\secref{subsec:trust-dynamics}.
%\todo{I don't get this part} Second, although the trust-POMDP robot policy and \baseline robot policy are different, 
%participants' reported (mean) trust evolved in a similar way. 
%This supports our notion that optimizing team performance might not maximize
%trust. 
%On the other hand, maximizing trust alone is insufficient to optimize team performance.
%This significantly differs our work from previous works on studying 
%trust dynamics and maximizing
%human trust~\cite{xu2015optimo,wang2016trust}, which focuses on real-time trust inference 
%and adapting robot behaviors to maximize human trust,
%while trust-POMDP focuses on optimizing the team performance. 
Second,
there is a lack of significant difference in the \emph{average} trust evolution between the two conditions (~\figref{fig:performance_figs}, center column), especially given that fewer human interventions occurred under the trust-POMDP policy. This can be partially explained by a combination of averaging and nonlinear trust dynamics, specifically that robot performance in the earlier part of the task has a more pronounced impact on trust~\citep{desai2012modeling}. 
This is a specific manifestation of the ``primacy effect'', a cognitive bias that results in a subject crediting a performer more if the performer succeeds earlier in time~\citep{jones1968pattern}. 
\figref{fig:timeEffectonTrust} shows this time-dependent aspect of trust dynamics in our experiment; the change in the mean of trust was larger if the robot succeeded earlier, most clearly seen for the Can and Glass objects in the real-robot experiment. 
As such, in the \baseline condition, although there were more interventions on the glass/can at the beginning, this was averaged out by a larger increase in the human trust. 

%why the time-dependent nonlinear effects have a stronger impact on high risk objects is beyond the scope of this work, we look forward to exploring this in the future.

% Link to psychology studies
%In psychology, this is known as 
% Link to Mujal's work

%The plots indicate that mean trust increased more if robot 
%succeeded earlier.
%In addition, the time-dependent nonlinear effect has a stronger impact on \emph{risker} objects, \ie,
%Can and Glass.
%Putting these all together explains~\figref{fig:performance_figs} (center column).
%In the \baseline condition, although there were more interventions on the glass/can
%at the beginning, this was balanced out by the fact that human trust increased 
%a lot if the robot succeeded earlier in time.
%In the trust-POMDP condition, although there were less interventions on the glass/can at the
%end, this was counterbalanced by the fact that human trust did not increase so much if the robot succeeded later in time.
%Finally, as to why the time-dependent nonlinear effects have a stronger impact on high risk objects is beyond
%the scope of this work, we look forward to exploring this in the future.

%\vspace{-0.1cm}
\subsection{Human behavioral policy}
\label{subsec:humanpolicy}

\figref{fig:performance_figs} (right column) 
shows the observed human behaviors given different trust levels. Consistent with the trust-based human behavioral model (\secref{subsec:human-behavioral-prediction}), participants were less likely to intervene as their trust in
the robot increased. The human's action also depended on the type of object. For low risk objects (bottles), participants allowed the robot's attempt to complete the task even if their trust in the robot was low. However, for a high risk object (glass), participants intervened unless they trusted the robot more.

%In summary, the trust-POMDP robot was able to infer participant's current trust level, and make the right decision as to whether to pick up the low risk object to build human trust, or to go directly to the high risk object when trust is high enough. This is the primary advantage that trust-POMDP robot has over the \baseline robot.

\begin{figure*}[t!]
\setlength
\tabcolsep{0pt}
\captionsetup[subfigure]{labelformat=empty}
\captionsetup[subfigure]{width=0.68\columnwidth}
\captionsetup[subfigure]{justification=justified,singlelinecheck=false}
\centering
\begin{tabular}{ccccc}
\begin{subfigure}[b]{.20\linewidth}
\centering
  \begin{tabular}{c}
      \includegraphics[width=0.9\linewidth]{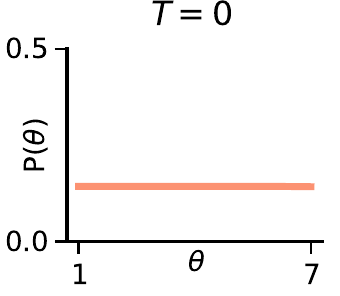}\\
  \includegraphics[width=0.85\linewidth]{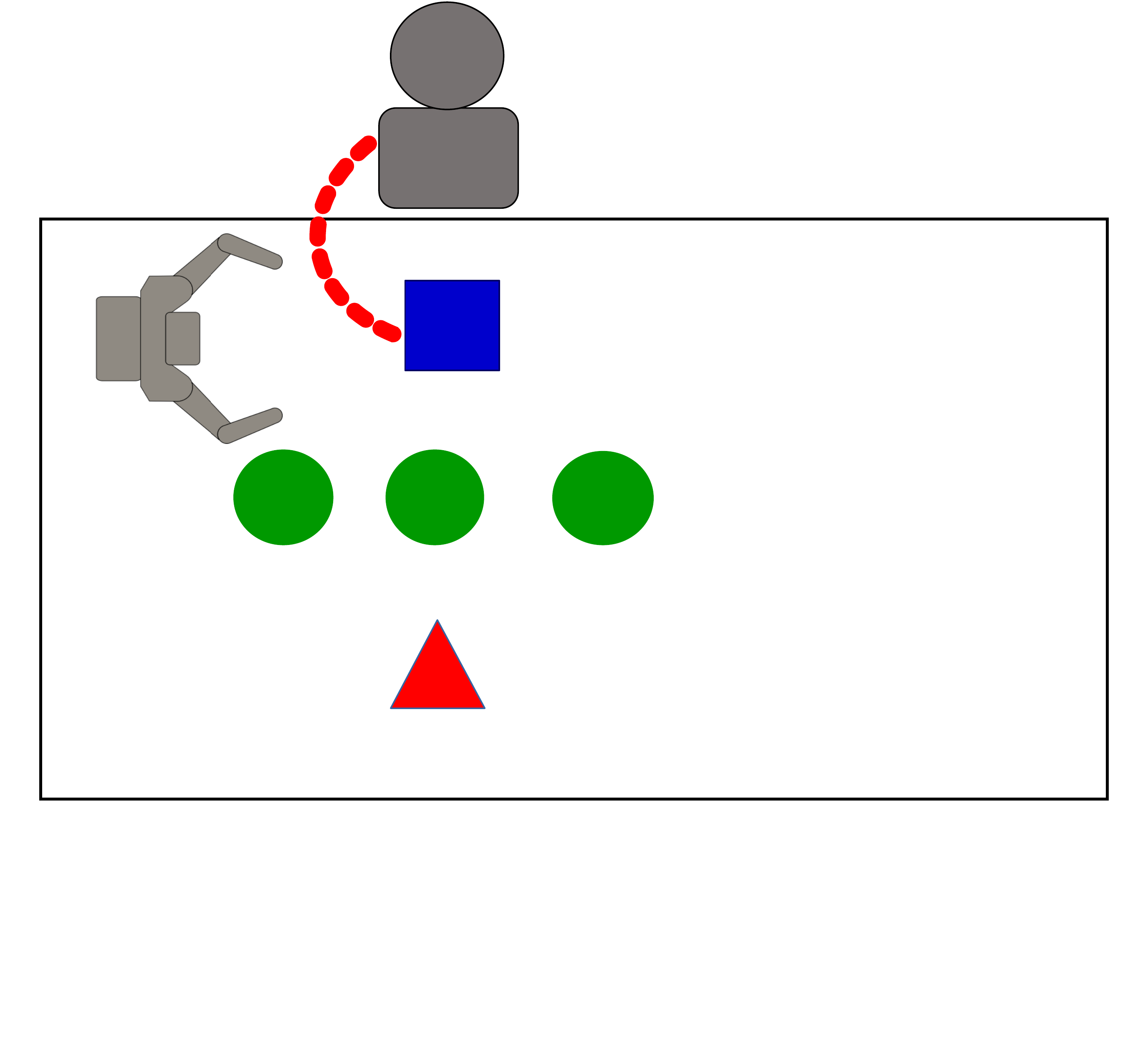}
     \end{tabular}
   \label{fig:plotT0}
\end{subfigure}
&
\begin{subfigure}[b]{.20\linewidth}
\centering
  \begin{tabular}{c}
      \includegraphics[width=0.9\linewidth]{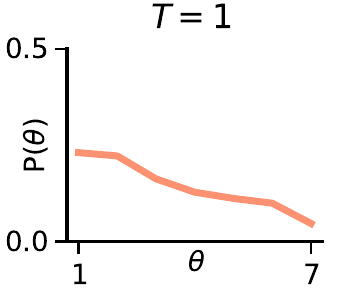}\\
  \includegraphics[width=0.85\linewidth]{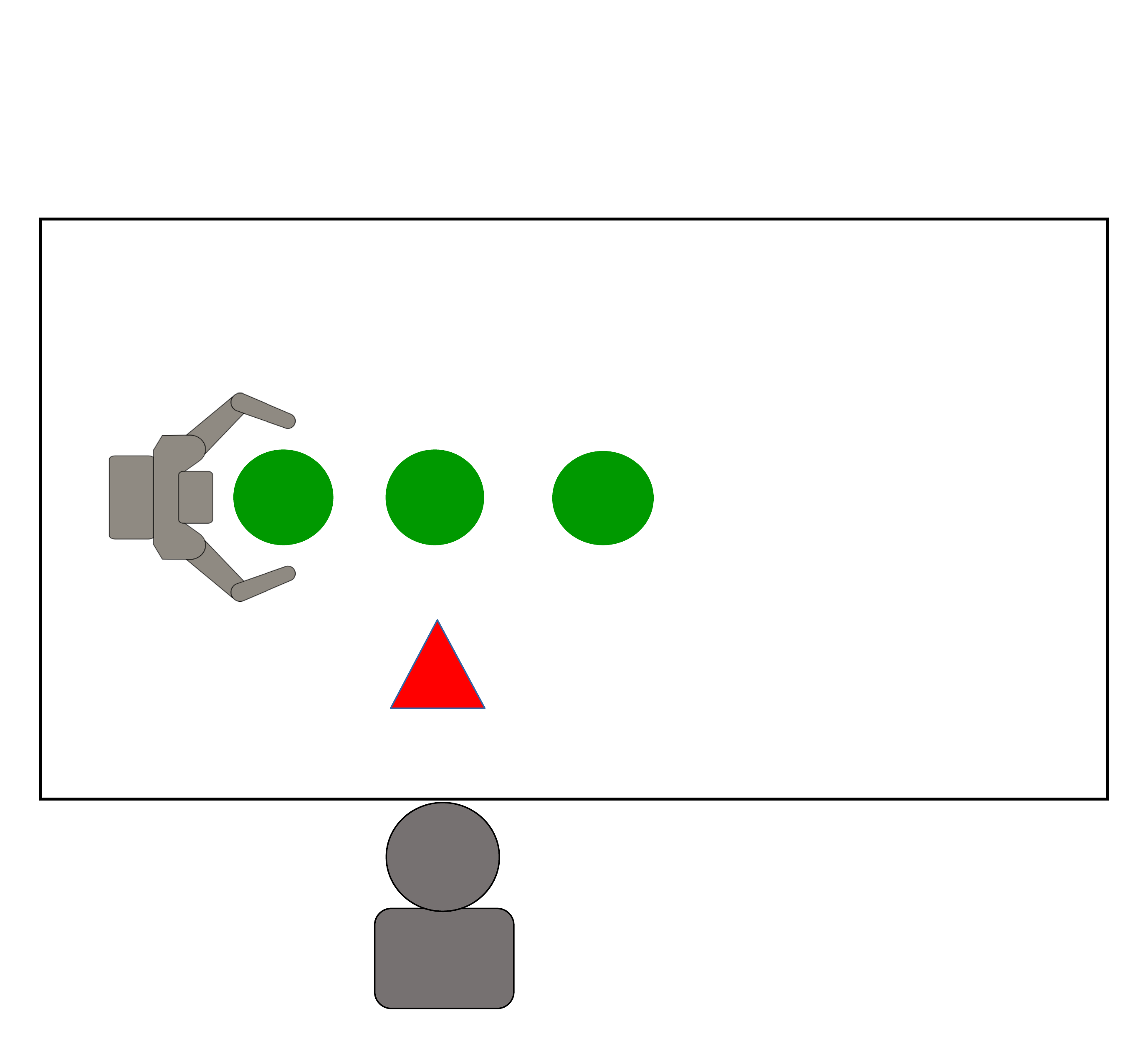}
   \end{tabular}
   \label{fig:plotT1}
\end{subfigure}
&
\begin{subfigure}[b]{.20\linewidth}
\centering
  \begin{tabular}{c}
  \fontsize{9}{5}\selectfont
  %\textbf{Trust-POMDP}\par\medskip
      \includegraphics[width=0.9\linewidth]{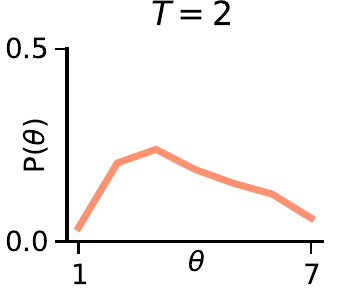}\\
  \includegraphics[width=0.85\linewidth]{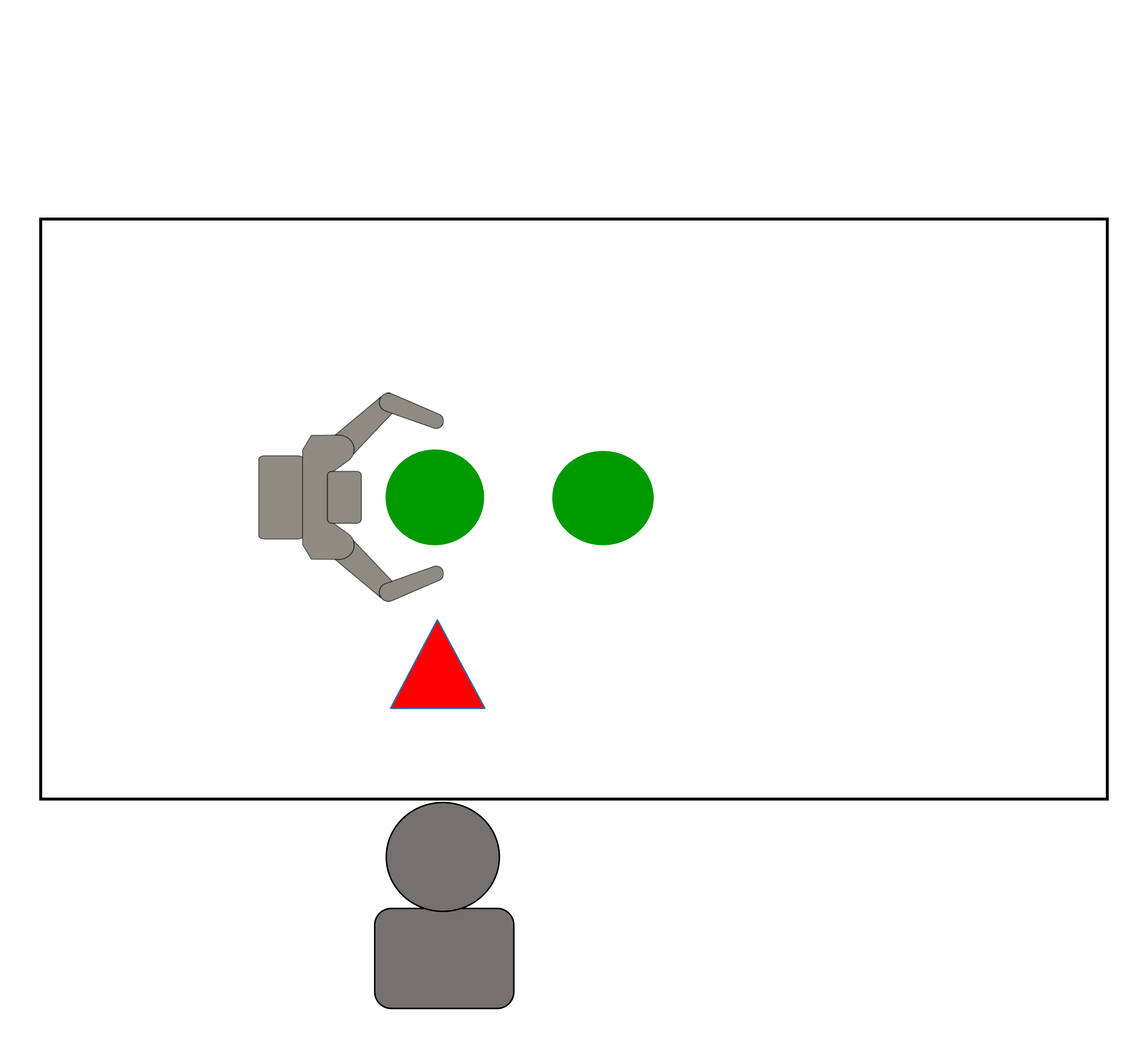}
   \end{tabular}
   \label{fig:plotT2}
\end{subfigure}
&
\begin{subfigure}[b]{.20\linewidth}
\centering
  \begin{tabular}{c}
      \includegraphics[width=0.9\linewidth]{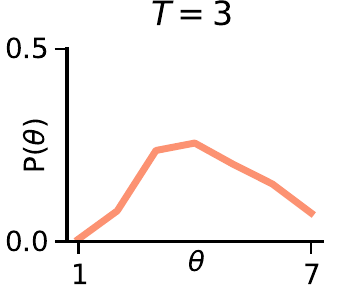}\\
  \includegraphics[width=0.85\linewidth]{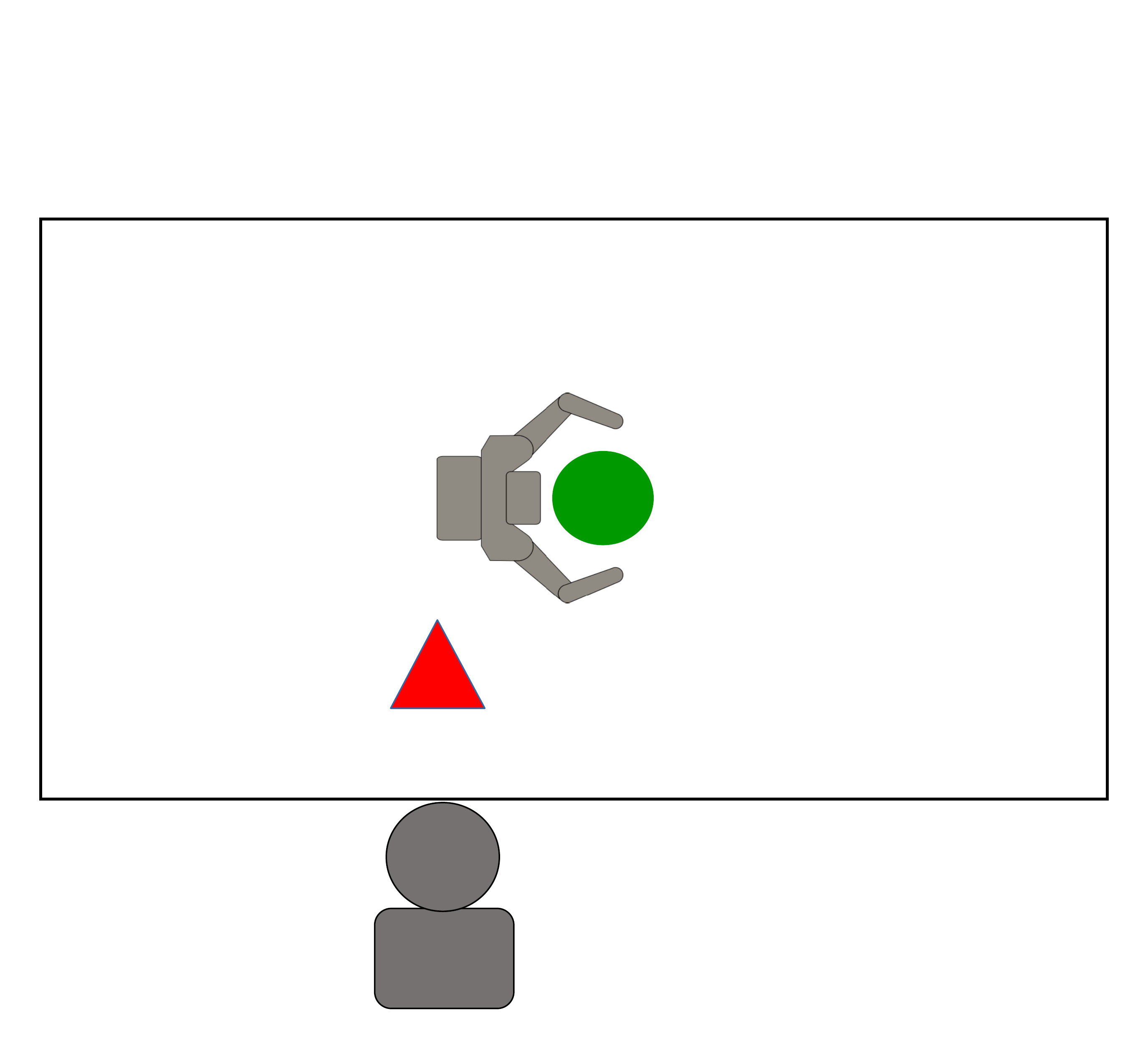}
     \end{tabular}
   \label{fig:plotT3}
\end{subfigure}
&
\begin{subfigure}[b]{.20\linewidth}
\centering
  \begin{tabular}{c}
  \includegraphics[width=0.9\linewidth]{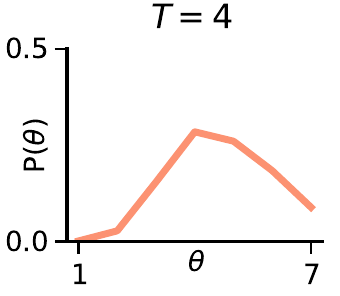}\\
  \includegraphics[width=0.85\linewidth]{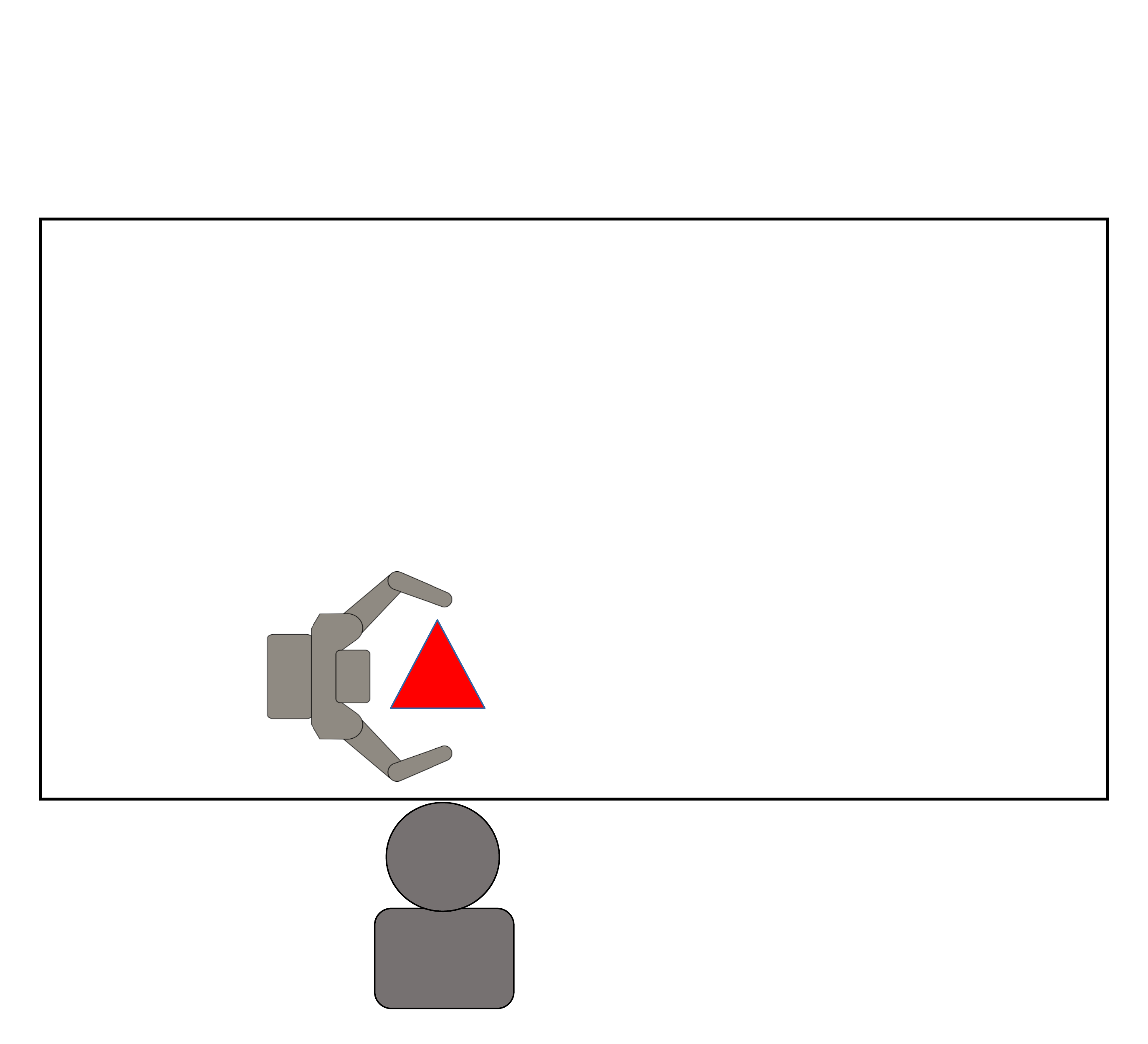}
   \end{tabular}
   \label{fig:plotT4}
\end{subfigure}
\end{tabular}
\caption{Sample run of the trust-POMDP strategy when the robot may fail in the glass cup with probability 0.9.}
\label{fig:fail-glass}
\end{figure*}

\begin{figure*}[t!]
\setlength
\tabcolsep{0pt}
\captionsetup[subfigure]{labelformat=empty}
\captionsetup[subfigure]{width=0.68\columnwidth}
\captionsetup[subfigure]{justification=justified,singlelinecheck=false}
\centering
\begin{tabular}{c}
\begin{tabular}{ccccc}
\begin{subfigure}[b]{.20\linewidth}
\centering
  \begin{tabular}{c}
      \includegraphics[width=0.9\linewidth]{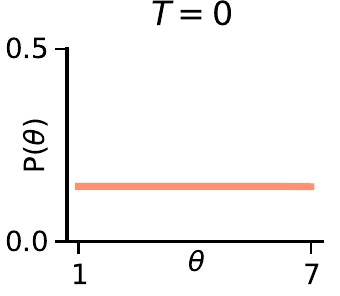}\\
  \includegraphics[width=0.85\linewidth]{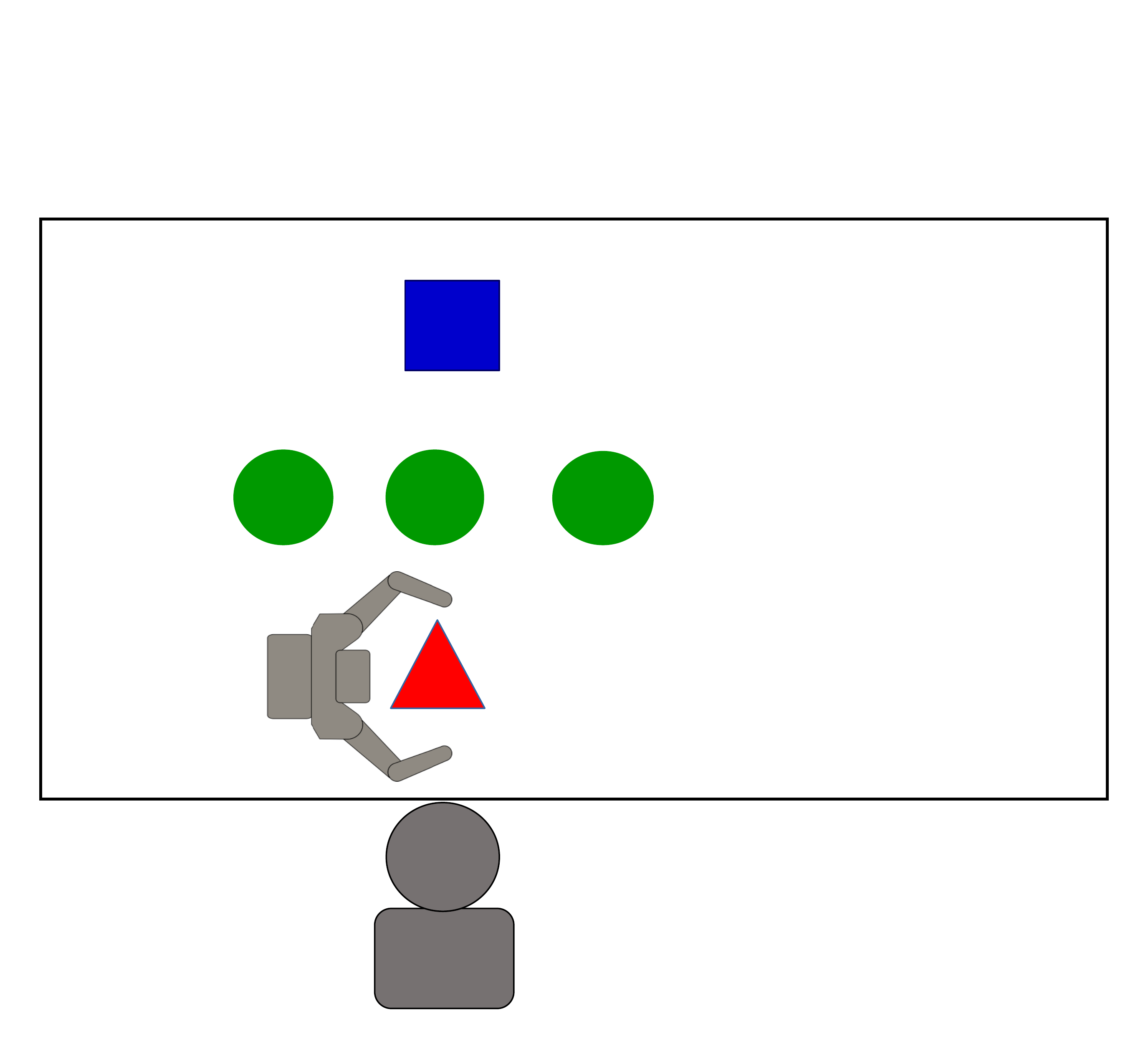}
     \end{tabular}
   \label{fig:plotT0}
\end{subfigure}
&
\begin{subfigure}[b]{.20\linewidth}
\centering
  \begin{tabular}{c}
      \includegraphics[width=0.9\linewidth]{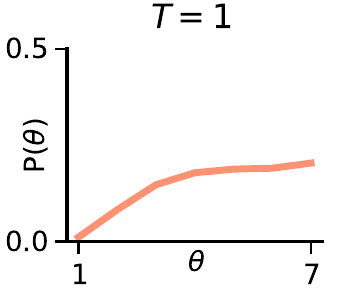}\\
  \includegraphics[width=0.85\linewidth]{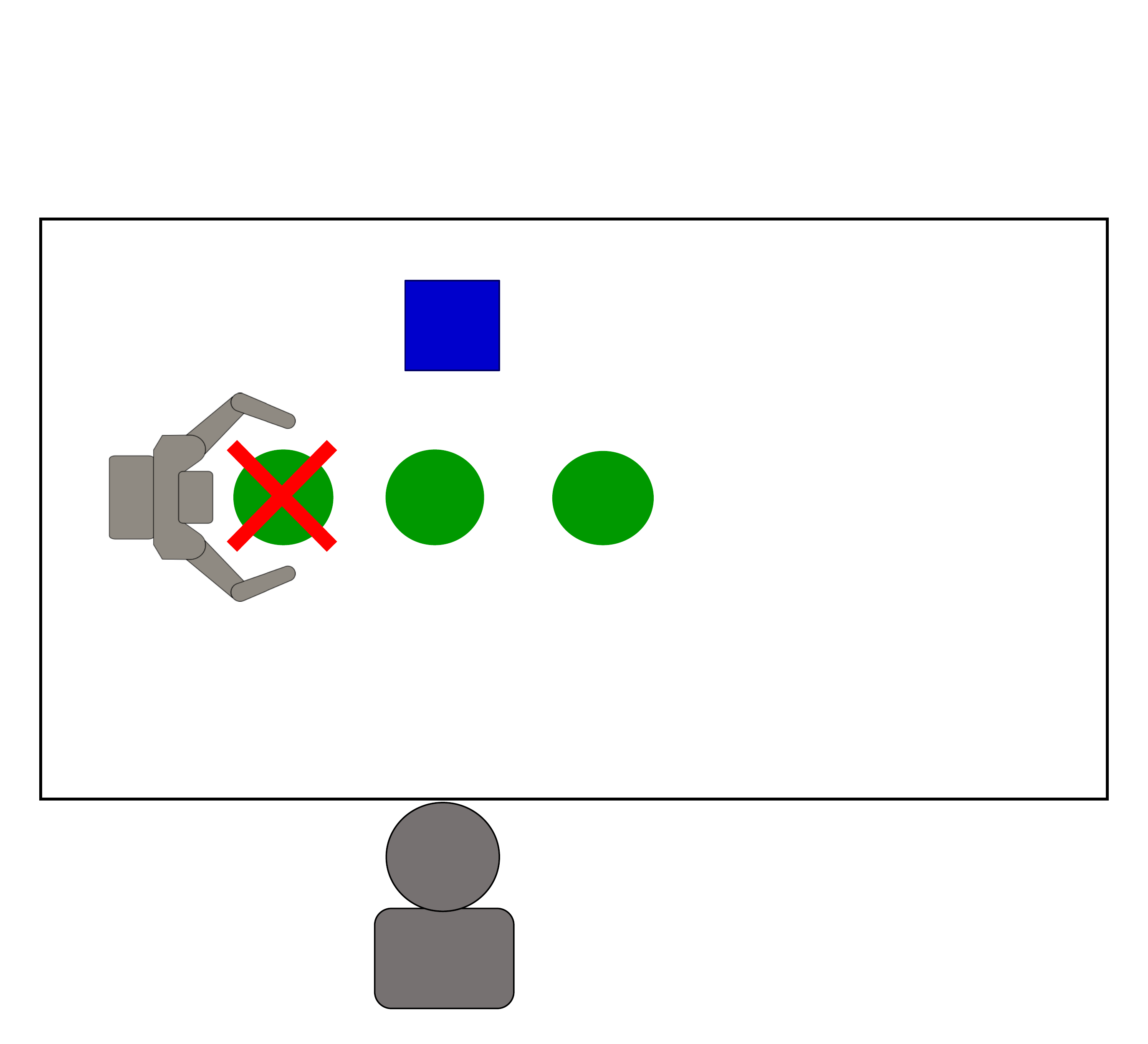}
   \end{tabular}
   \label{fig:plotT1}
\end{subfigure}
&
\begin{subfigure}[b]{.20\linewidth}
\centering
  \begin{tabular}{c}
  \fontsize{9}{5}\selectfont
  %\textbf{Trust-POMDP}\par\medskip
      \includegraphics[width=0.9\linewidth]{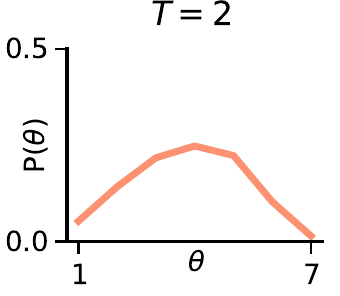}\\
  \includegraphics[width=0.85\linewidth]{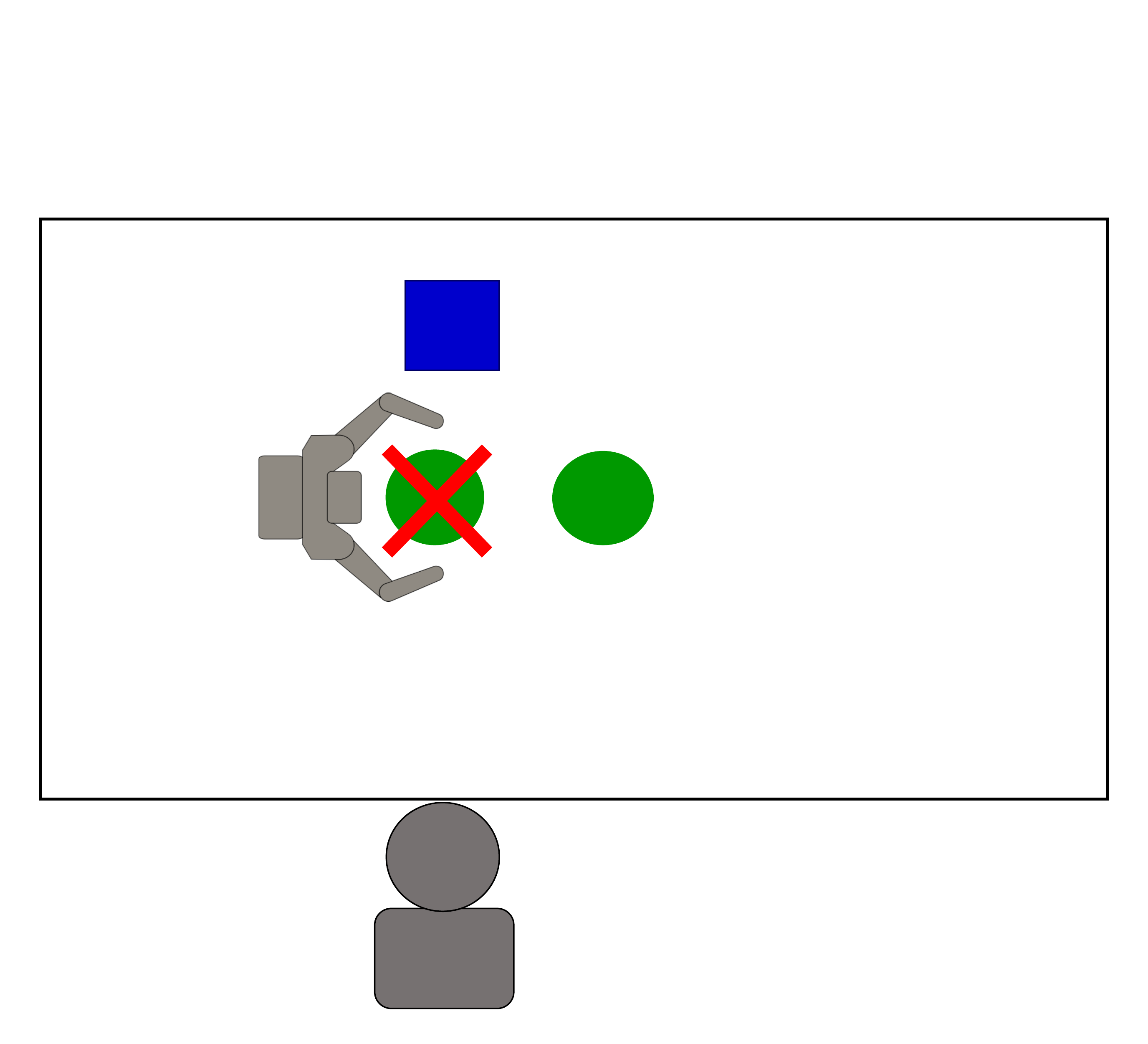}
   \end{tabular}
   \label{fig:plotT2}
\end{subfigure}
&
\begin{subfigure}[b]{.20\linewidth}
\centering
  \begin{tabular}{c}
      \includegraphics[width=0.9\linewidth]{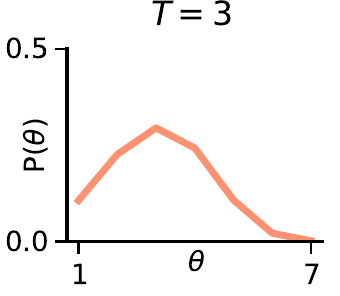}\\
  \includegraphics[width=0.85\linewidth]{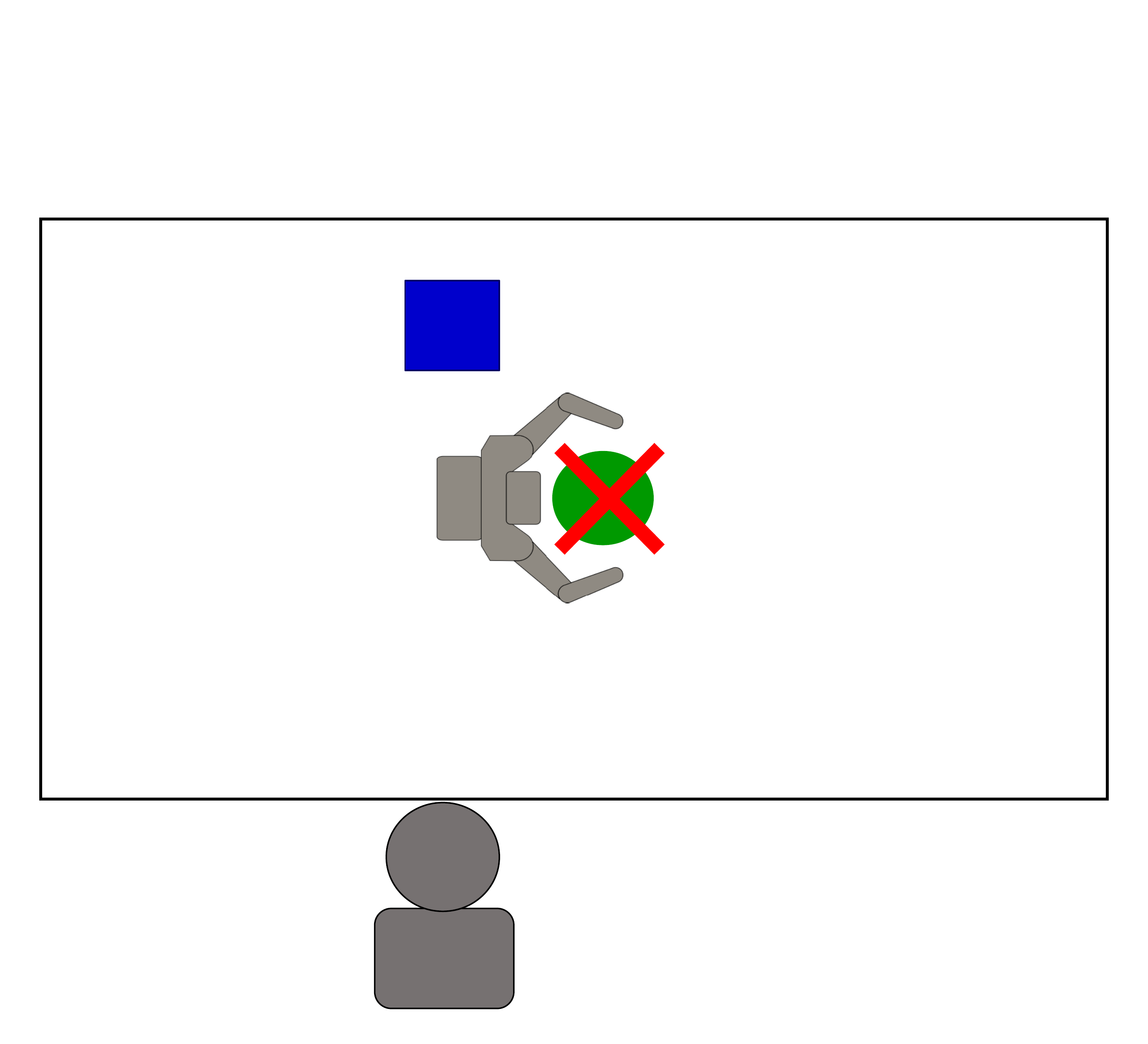}
     \end{tabular}
   \label{fig:plotT3}
\end{subfigure}
&
\begin{subfigure}[b]{.20\linewidth}
\centering
  \begin{tabular}{c}
  \includegraphics[width=0.9\linewidth]{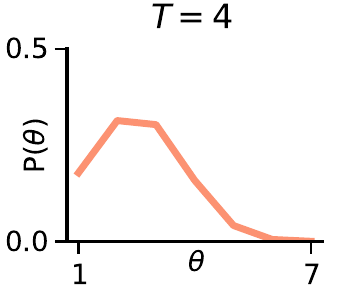}\\
  \includegraphics[width=0.85\linewidth]{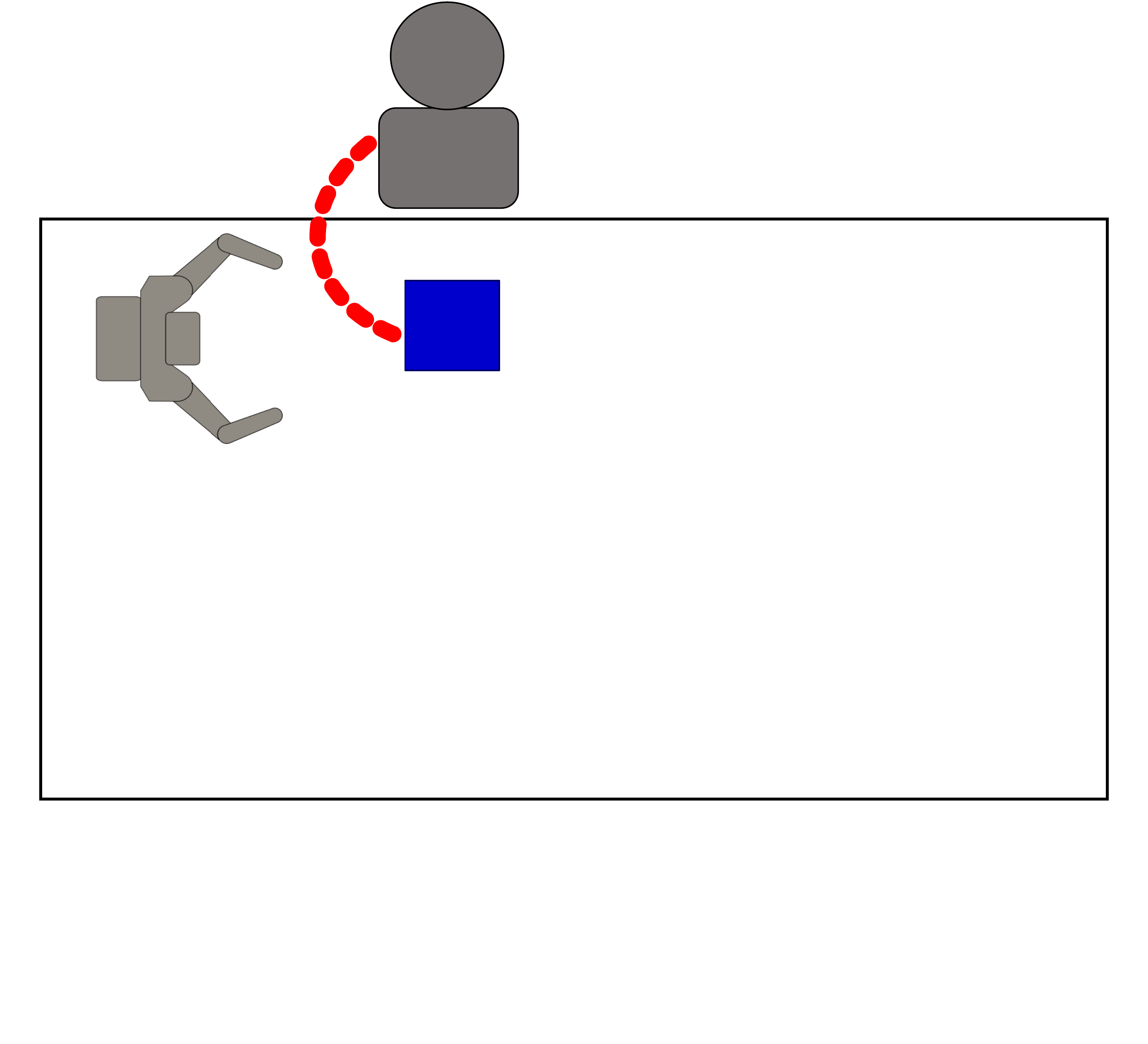}
   \end{tabular}
   \label{fig:plotT4}
\end{subfigure}
\end{tabular}~\\
\begin{tabular}{ccccc}
\begin{subfigure}[b]{.20\linewidth}
\centering
  \begin{tabular}{c}
      \includegraphics[width=0.9\linewidth]{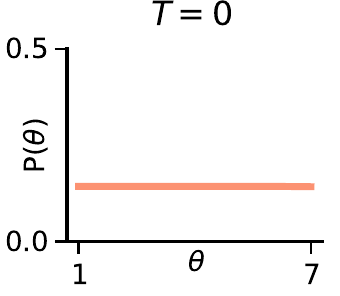}\\
  \includegraphics[width=0.85\linewidth]{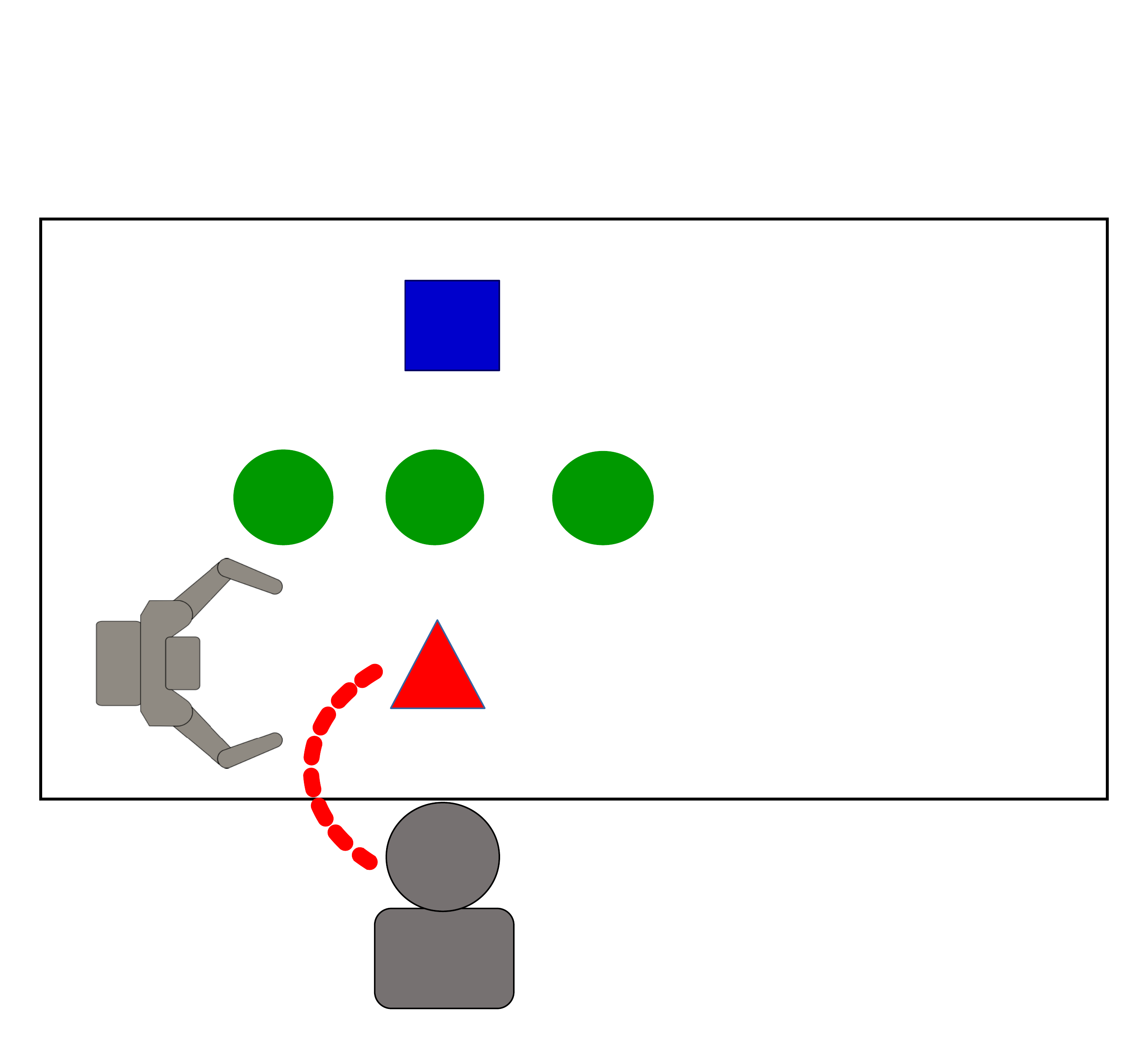}
     \end{tabular}
   \label{fig:plotT0}
\end{subfigure}
&
\begin{subfigure}[b]{.20\linewidth}
\centering
  \begin{tabular}{c}
      \includegraphics[width=0.9\linewidth]{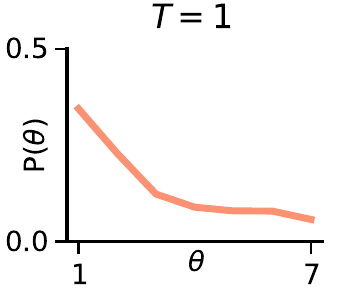}\\
  \includegraphics[width=0.85\linewidth]{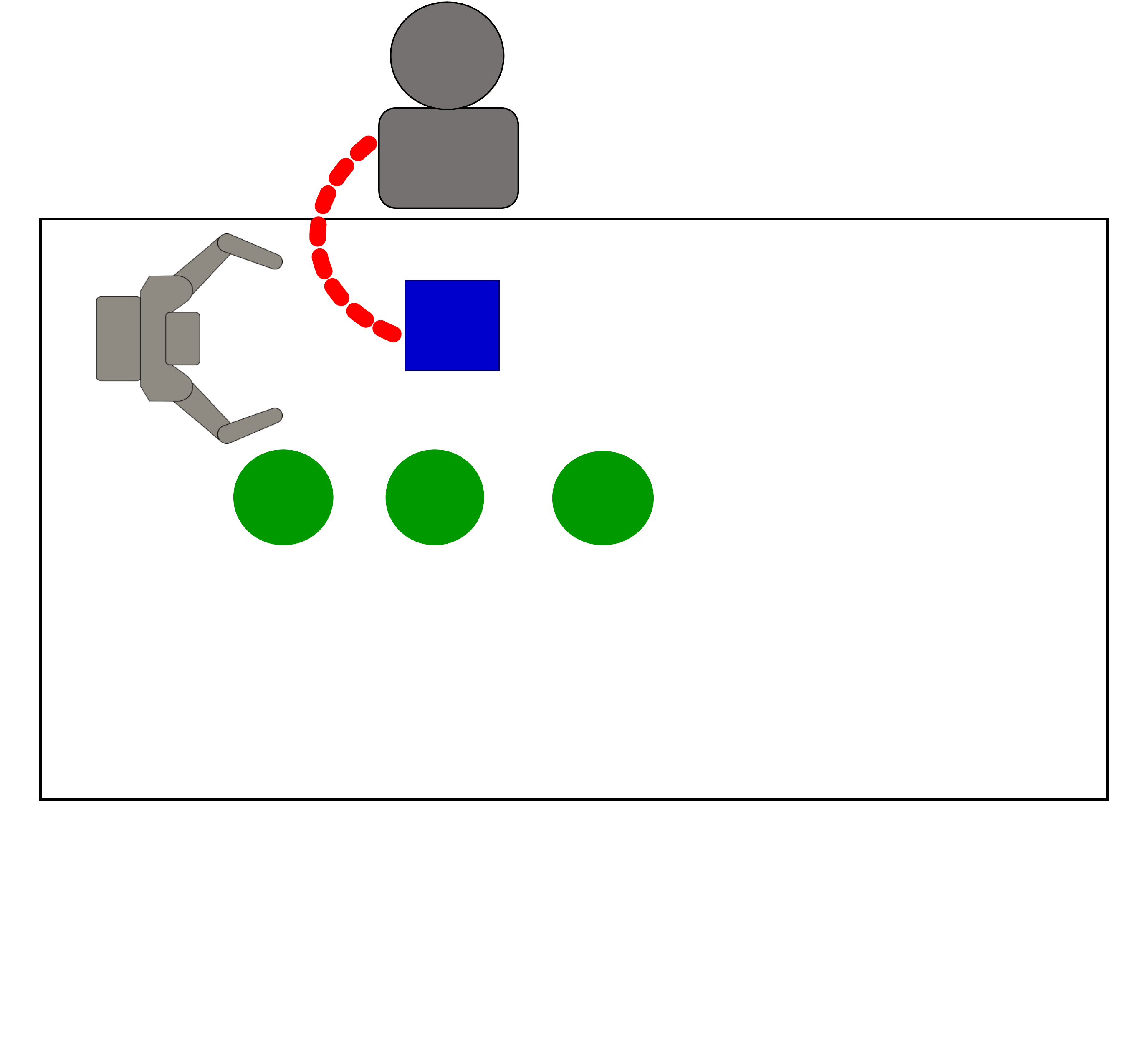}
   \end{tabular}
   \label{fig:plotT1}
\end{subfigure}
&
\begin{subfigure}[b]{.20\linewidth}
\centering
  \begin{tabular}{c}
  \fontsize{9}{5}\selectfont
  %\textbf{Trust-POMDP}\par\medskip
      \includegraphics[width=0.9\linewidth]{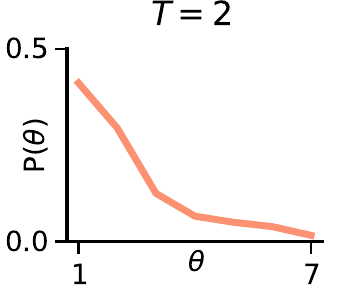}\\
  \includegraphics[width=0.85\linewidth]{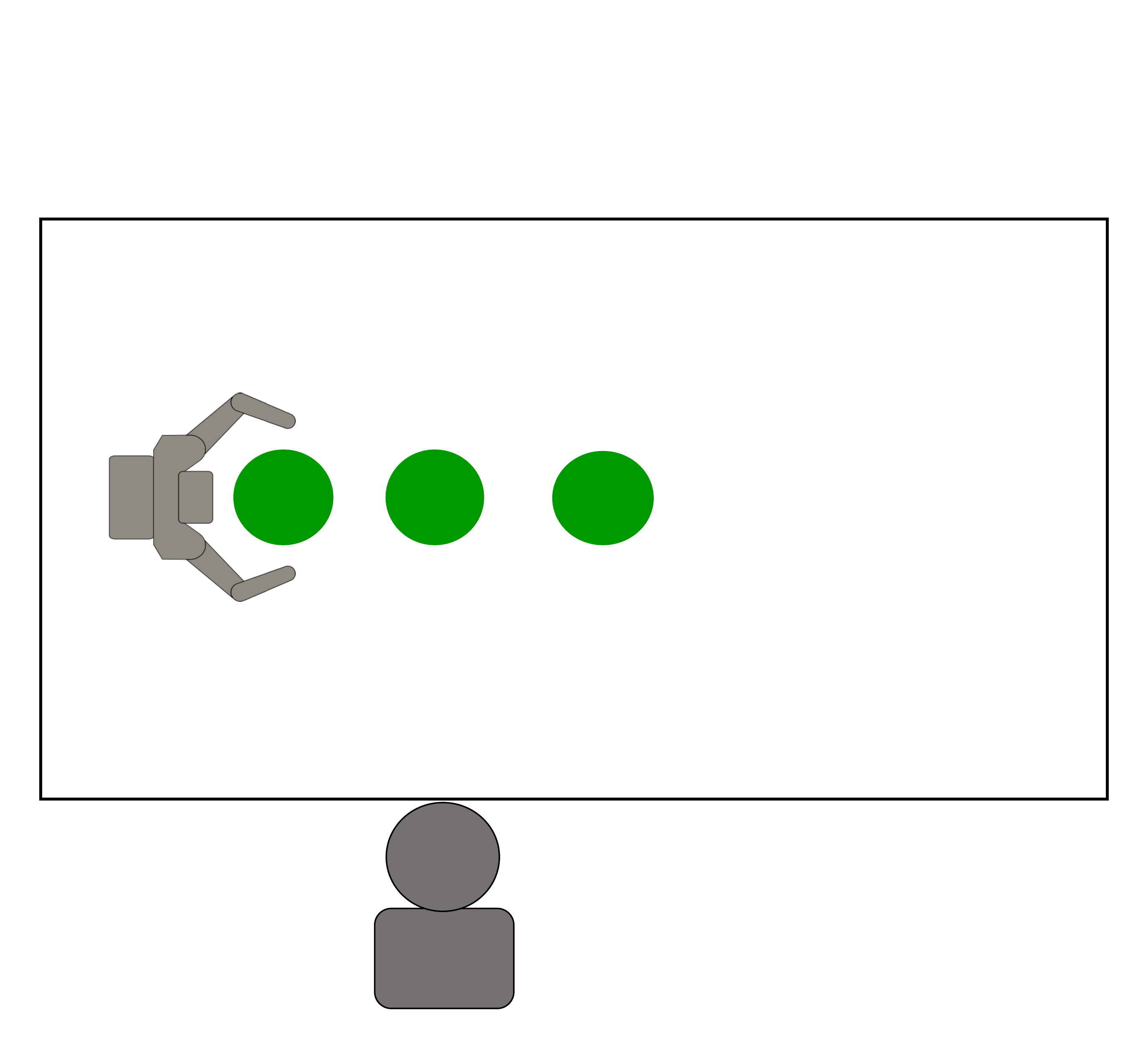}
   \end{tabular}
   \label{fig:plotT2}
\end{subfigure}
&
\begin{subfigure}[b]{.20\linewidth}
\centering
  \begin{tabular}{c}
      \includegraphics[width=0.9\linewidth]{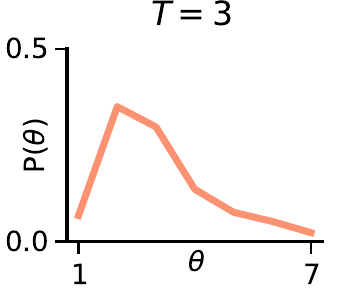}\\
  \includegraphics[width=0.85\linewidth]{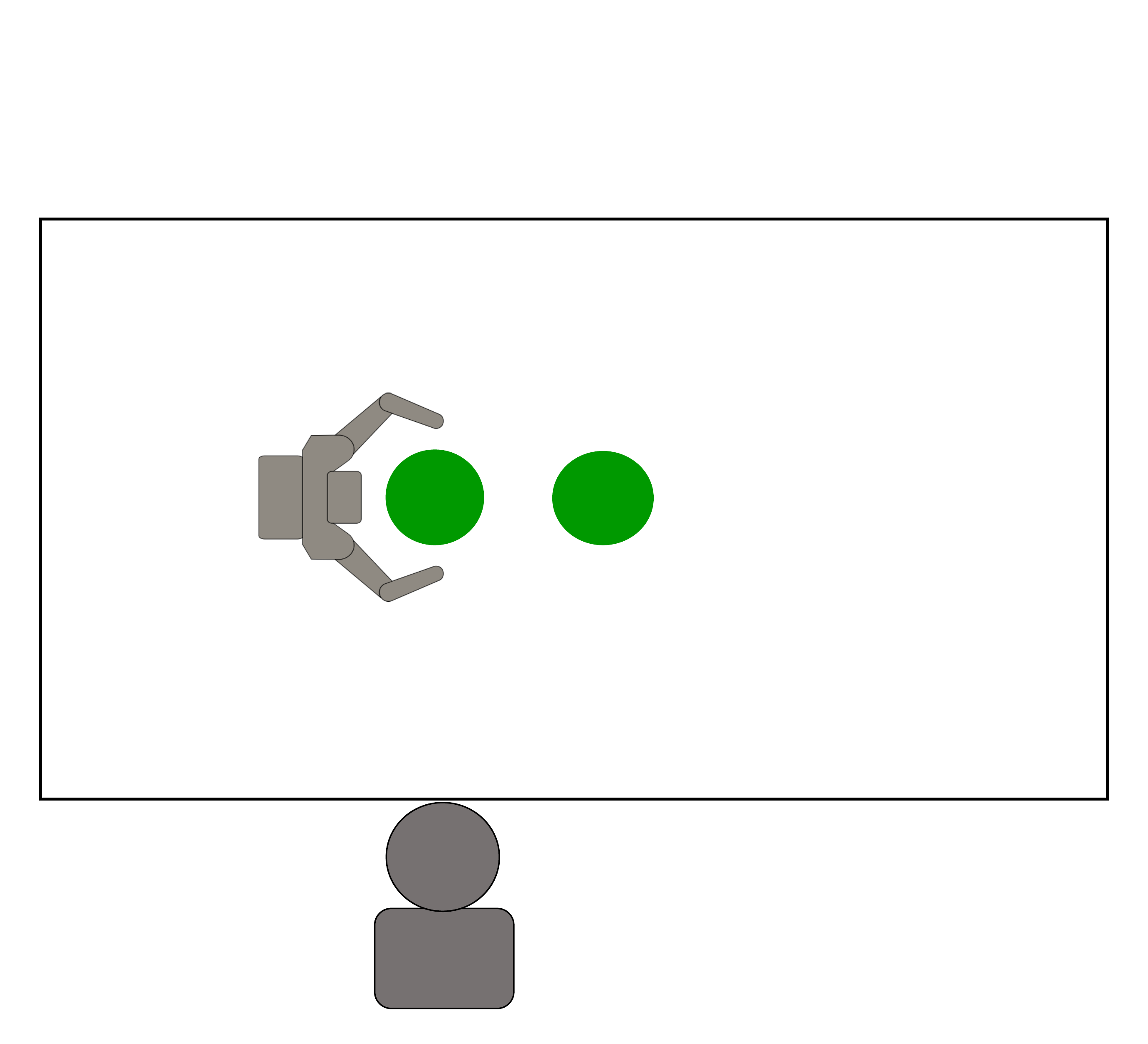}
     \end{tabular}
   \label{fig:plotT3}
\end{subfigure}
&
\begin{subfigure}[b]{.20\linewidth}
\centering
  \begin{tabular}{c}
  \includegraphics[width=0.9\linewidth]{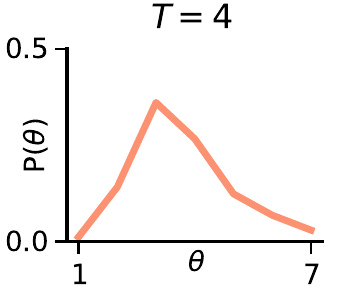}\\
  \includegraphics[width=0.85\linewidth]{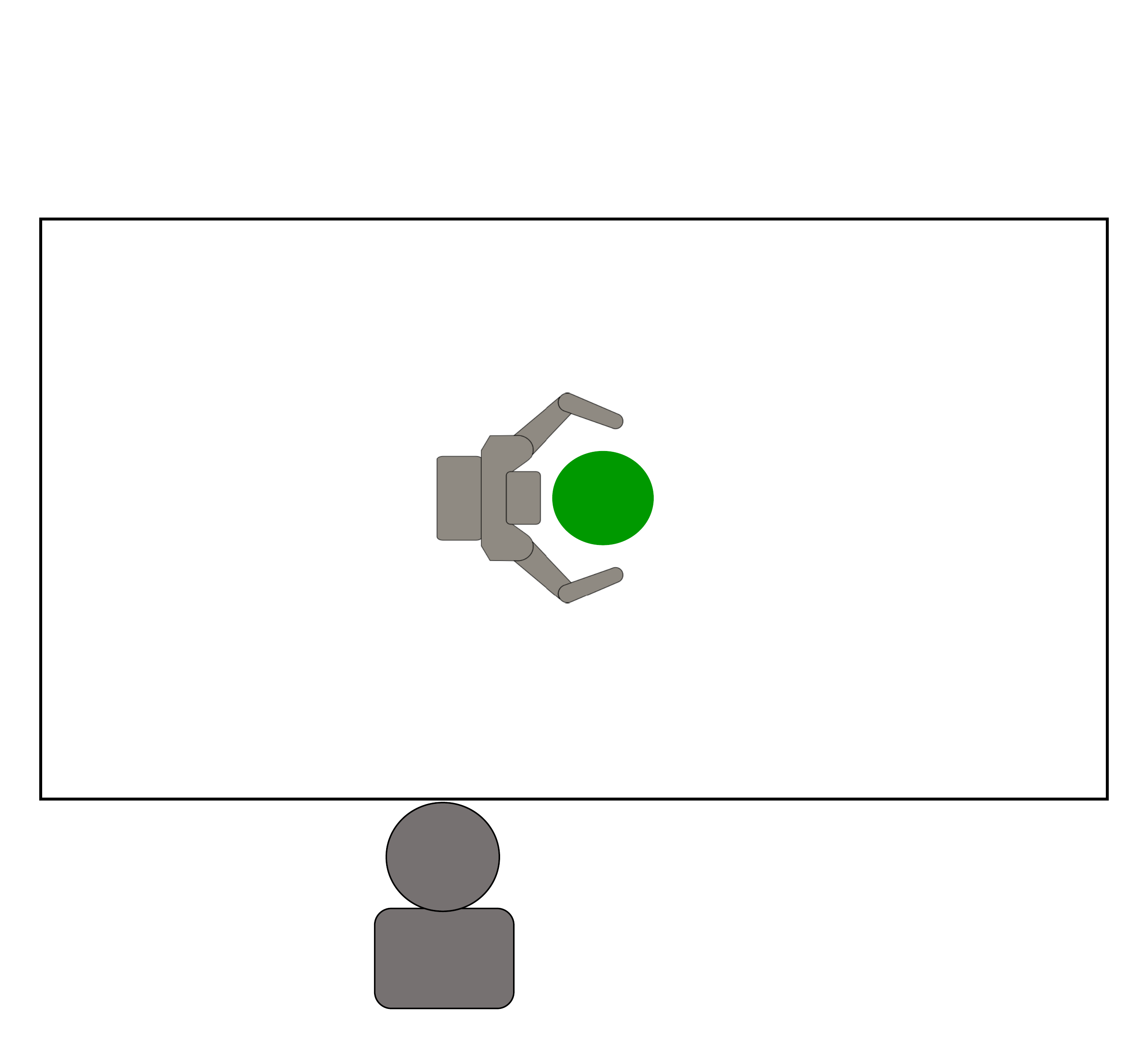}
   \end{tabular}
   \label{fig:plotT4}
\end{subfigure}
\end{tabular}~\\~\\~\\
\begin{tabular}{ccccc}
\begin{subfigure}[b]{.20\linewidth}
\centering
  \begin{tabular}{c}
  \includegraphics[width=0.85\linewidth]{failure-figs/glass-fail-01.pdf}
     \end{tabular}
   \label{fig:plotT0}
\end{subfigure}
&
\begin{subfigure}[b]{.20\linewidth}
\centering
  \begin{tabular}{c}
  \includegraphics[width=0.85\linewidth]{failure-figs/glass-fail-02.pdf}
   \end{tabular}
   \label{fig:plotT1}
\end{subfigure}
&
\begin{subfigure}[b]{.20\linewidth}
\centering
  \begin{tabular}{c}
  \fontsize{9}{5}\selectfont
  %\textbf{Trust-POMDP}\par\medskip
  \includegraphics[width=0.85\linewidth]{failure-figs/glass-fail-03.pdf}
   \end{tabular}
   \label{fig:plotT2}
\end{subfigure}
&
\begin{subfigure}[b]{.20\linewidth}
\centering
  \begin{tabular}{c}
  \includegraphics[width=0.85\linewidth]{failure-figs/glass-fail-04.pdf}
     \end{tabular}
   \label{fig:plotT3}
\end{subfigure}
&
\begin{subfigure}[b]{.20\linewidth}
\centering
  \begin{tabular}{c}
  \includegraphics[width=0.85\linewidth]{failure-figs/glass-fail-05.pdf}
   \end{tabular}
   \label{fig:plotT4}
\end{subfigure}
\end{tabular}
\end{tabular}
\caption{Sample runs of the performance-maximizing policy (top, middle-row) and the trust-maximizing policy (bottom row) when the robot may fail in the glass cup with probability 0.9, and the robot can fail intentionally in any object. The adaptive trust-POMDP policy branches out at $T=0$: If the human stays put (top row), the robot intentionally fails in the bottles to reduce human trust and maximize the probability of the human intervening when it goes for the glass at $T=4$.}
\label{fig:fail-intent}
\end{figure*}

\begin{figure}[t!]
\setlength
\tabcolsep{1pt}
%\captionsetup[subfigure]{labelformat=empty}
%\captionsetup[subfigure]{width=1.0\columnwidth}
\captionsetup[subfigure]{justification=justified,singlelinecheck=false}
\centering
\begin{subfigure}[b]{\linewidth}
\begin{tabular}{cc}
\begin{subfigure}[b]{.5\linewidth}
\centering
\includegraphics[width=0.9\linewidth]{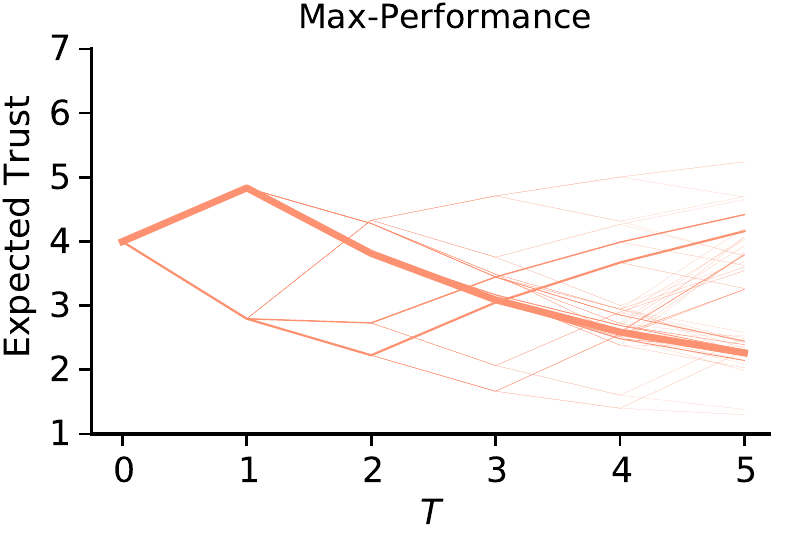}
\end{subfigure}
& 
\begin{subfigure}[b]{.5\linewidth}
\centering
\includegraphics[width=0.9\linewidth]{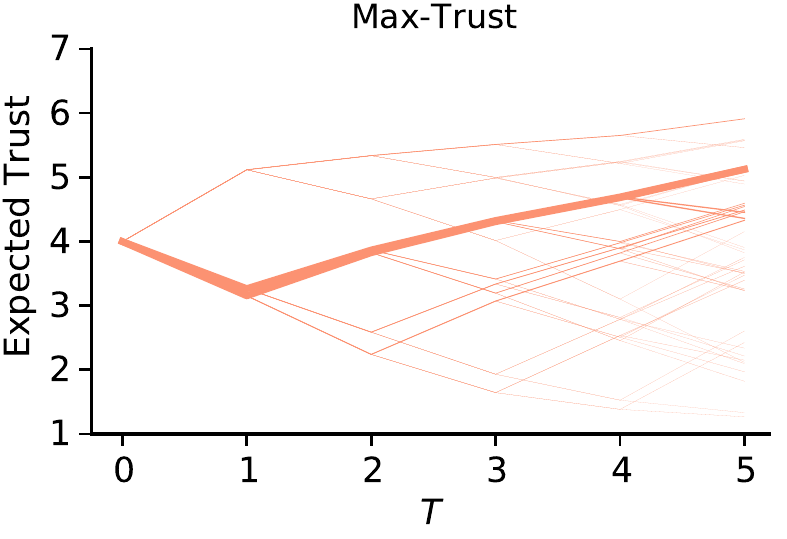}     \
\end{subfigure}
\end{tabular}
\end{subfigure}
\begin{subfigure}[b]{\linewidth}
\begin{tabular}{cc}
\begin{subfigure}[b]{.5\linewidth}
\centering
\includegraphics[width=0.9\linewidth]{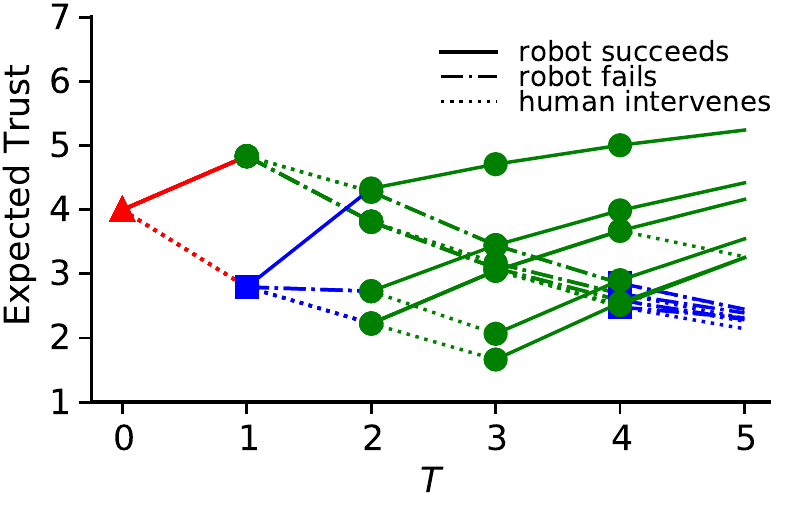}
\end{subfigure}
& 
\begin{subfigure}[b]{.5\linewidth}
\centering
\includegraphics[width=0.9\linewidth]{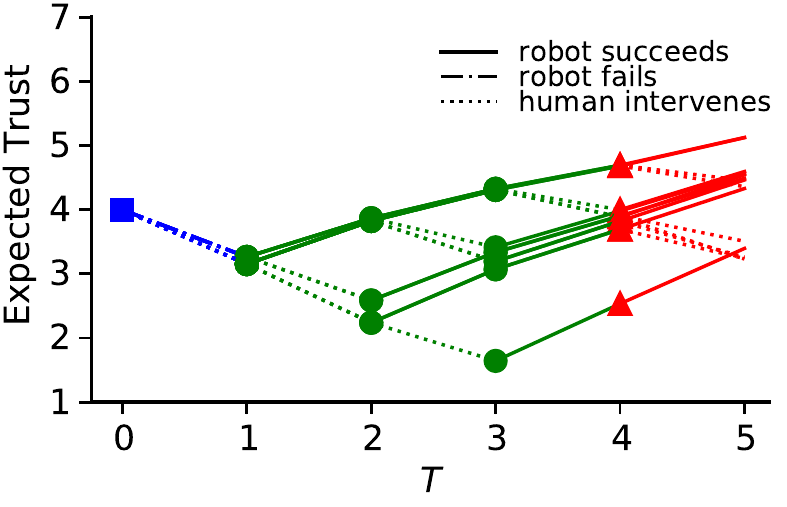}     \
\end{subfigure}
\end{tabular}
\end{subfigure}
\caption{(Top) Expected trust for all possible human action sequences for the performance-maximizing and trust-maximizing policy. Each sequence is represented with a line of width proportional to the likelihood of that sequence, based on the learned model. (Bottom) Annotated robot actions for the 16 most likely sequences.}
\label{fig:policy-branches}
\end{figure}

\begin{figure*}[t!]
\includegraphics[width=1.0\linewidth]{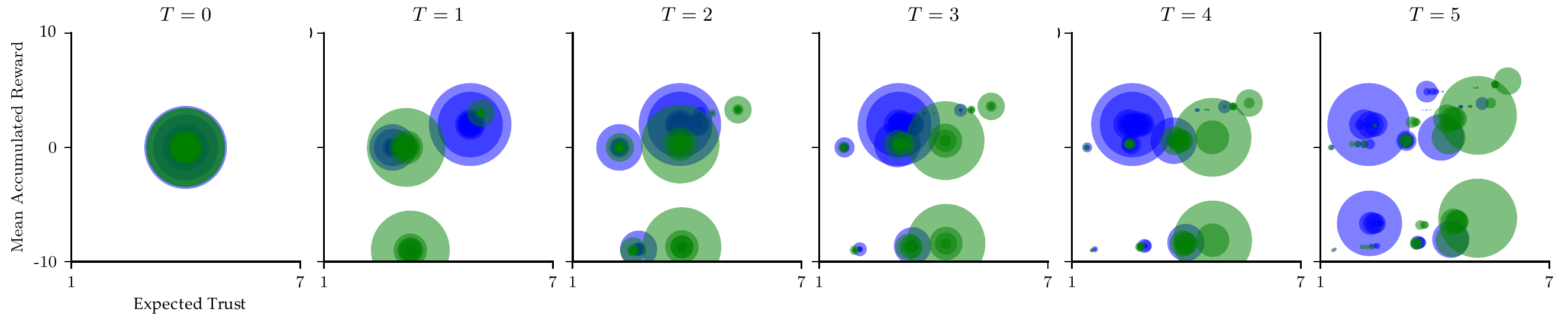} 
\caption{Scatterplot of mean accumulated reward as a function of human trust over time for all human action sequences. The radius of each circle is proportional to the likelihood of the corresponding sequence, based on the learned model. The performance-maximizing policy (blue) gradually reduces human trust to maximize the accumulated reward, while the trust-maximizing policy (green) focuses on increasing trust.}
\label{fig:film-strip}
\end{figure*} 

\section{Robot Failures}
The previous experimental results show that the trust-POMDP policy significantly outperforms the myopic policy that ignores trust in robot decision-making. The trust-POMDP robot was able to make good decisions on whether to pick up the low risk object to increase human trust, or to go directly to the high risk object when trust is high enough. This is one main advantage that trust-POMDP robot has over the \baseline robot.

In these experiments the robot always succeeded. However, in the real world the robot is also likely to fail, and we want to explore the behavior of the trust-POMDP when the robot may fail in its attempt to pick up an object with some known probability. 

Therefore, we assumed that the robot may fail when attempting to pick up the glass with probability 0.9, and we used the learned dynamics and human behavioral model to compute the robot policy in that case. Contrary to when the robot always succeeds, in this case it is actually beneficial for the human to intervene and pick up the glass themselves, in order to avoid the large penalty from a likely robot failure. 

Fig.~\ref{fig:fail-glass} shows the computed policy and belief updates: the robot starts with the glass cup, since the beginning of the task is when the human is the most likely to intervene and not let the robot pick up the glass (and likely fail in the process of doing so). 

While this shows that the robot can reason over human intervention rate to reduce failure, intuitively the robot should also be able to \emph{actively reduce} trust to affect human behavior. While there is a range of behaviors that can reduce human trust~\cite{wang2016trust,van2014robot}, we focused on active trust reduction through failures. Therefore, we expanded the robot's action space, so that it can intentionally fail in any object. Keeping the failure probability for glass at 0.9 and reducing the reward for robot success when picking up the bottles to 0.3 results in the exciting behavior demonstrated at Fig.~\ref{fig:fail-intent}. 

When following the trust-POMDP policy (Fig.~\ref{fig:fail-intent} top and middle row) the robot attempts to pick up the can first; This is an \emph{information seeking} action, that the robot uses to estimate the initial human trust. If the human stays put, the robot infers that human trust is high, and it will then \emph{fail intentionally} at the bottles to reduce trust, before going for the glass cup. By the time the robot  goes for the glass cup, human trust has been reduced sufficiently so that the human is likely to intervene, avoiding failure. On the other hand, if the human intervenes, the robot infers that the human trust is already low. The robot then does not need to fail intentionally, since it does not need to reduce human trust any further, but it subsequently goes for the glass cup.

The resulting policy contrasts the policy that the robot follows, if it maximizes human trust instead (Fig.~\ref{fig:fail-intent}, bottom row). When following the trust-maximizing policy, the robot starts with the glass. This is for two reasons: (a)  in the beginning human trust is the lowest, therefore the human is the most likely to intervene and avoid watching the robot fail, which would result in significant reduction in trust (b) Even if the human does not intervene and the robot fails, it is better to fail early when trust has not increased yet, since the higher the trust, the steeper the fall, based on the learned model of Fig.~\ref{fig:trustdyna}.

We further illustrate the difference between the two policies by simulating policy runs and showing the evolution of the expected trust and mean accumulated reward over time (Fig.~\ref{fig:policy-branches}, ~\ref{fig:film-strip}). The plots illustrate how the performance-maximizing policy reduces human trust to maximize reward. The mean accumulated reward over $10^{4}$ policy runs for the performance-maximizing policy is $-1.36$, compared to $-1.65$ for the trust-maximizing policy, a statistically significant difference ($F(1,19998) = 18.04, p < 0.001$). This evaluation indicates that maximizing trust can be suboptimal in the presence of robotic failures.

\section{Conclusion} 
\label{sec:Conclusion}
%The experimental results show that the trust-POMDP policy significantly outperforms the myopic policy that ignores trust in robot decision-making. The trust-POMDP robot was able to make good decisions on whether to pick up the low risk object to increase human trust, or to go directly to the high risk object when trust is high enough. This is one main advantage that trust-POMDP robot has over the \baseline robot.

This paper presents the trust-POMDP, a computational model for integrating
human trust into robot decision making.  The trust-POMDP closes the loop
between trust models and robot decision making.  It enables the robot to
infer and influence human trust systematically and to leverage trust for fluid collaboration.

Our experimental results in a table-clearing task show that the trust-POMDP policy calibrates human trust to match it to the robot's manipulation capabilities: If trust is overly low, the robot prioritizes picking up the low risk objects to increase trust. This results in better performance, compared to the \baseline robot that ignores trust. On the other hand, if trust is overly high, the robot fails intentionally in the low risk objects. Our results show that always maximizing trust can be in fact detrimental to performance in the presence of robotic failures. 

%The differs from approaches that focus on on trust maximization~\citep{xu2015optimo,wang2016trust}, \ie, the trust-POMDP uses trust as the means to maximize task performance and builds up human trust only when necessary. 

% === Remove the details of the Tesla accident ===
%For example, a
%Tesla Model S vehicle with its autopilot activated was involved in a fatal
%accident in $2016$.
%The accident occurred at an intersection, when a
%tractor trailer making a turn crossed into the path of the Model S.  Both the
%Model S' driver, who, according to Tesla, is ultimately responsible for the
%vehicle's actions, and the autopilot system failed to detect the big trailer,
%and did not apply the brakes.  Tesla has been consistently telling its
%customers that the autopilot is in a "public beta" phase and that the system
%is designed with the expectation that drivers keep their hands on the wheel.
%However, at the time of the accident, the autopilot had been performing well for
%over 130 million miles, and the driver became over-reliant on the autopilot
%and lost situation awareness.  
%We believe  that by modeling human trust
%and connecting it with human and robot decision-making, the trust-POMDP will
%be able to detect human over-reliance on the robot and intervene to engage the
%human when necessary.  We are excited to explore this direction in  future work.

% Limitations
% \subsection{Limitations}
There are several limitations in our current work. Similar to previous works~\citep{xu2015optimo,desai2012modeling}, we modeled trust as a single real-valued latent variable that reflected the capabilities of the entire system. However, a multi-dimensional parameterization of trust that captured the different functions and modes of automation could be be a more accurate representation. In addition, the evolution of trust might also depend on the type of motion executed by the robot (e.g., for expressive or deceptive motions~\citep{dragan2013legibility,dragan2014analysis}). The current trust-POMDP model also assumes static robot capabilities, but a robot's true capabilities may change over time. In fact, the trust-POMDP can be extended to model robot capabilities via additional
state variables that affect the state transition dynamics. Furthermore, the reward function is manually specified in this work. However, the reward function may be difficult to specify in practice. One possible way to resolve this is to learn the reward function from human demonstrations (e.g., ~\citep{nikolaidis2015efficient}). Finally, the trust model learned on one task may transfer to a related task~\cite{soh2018transfer}. This last aspect is another interesting direction for future work.

% I'm not sure as to the point of this section? We have already mentioned "trust calibration" before. The task transfer is not really related to the issue of model-free learning. The latter can potentially generalize to new tasks/robots as well. 

%\vspace{-0.05cm}
%\section{Conclusion}
%\label{sec:conclusion}

%We compared the trust-POMDP strategy with a baseline myopic strategy, and obtained promising results. 
% This work opens up exciting  new opportunities for future work, including, in particular, modulating human trust to avoid over-reliance on robot autonomy and trust transfer across related task domains. 

% \noindent\textbf{Future work.}
%This work points to several directions for further investigation. In addition to addressing the  limitations aforementioned, we consider two particularly promising topics. First, over-trust may sometimes degrade task performance, as it causes the human to over-rely on robot autonomy. We believe that by modeling human trust and connecting it with human and robot decision-making, the trust-POMDP will be able to detect human over-reliance and intervene to engage the human when necessary. Second, the trust-POMDP is a generative decision model conditioned explicitly on trust. Explicit trust modeling provides several advantages: it fits better to experimental data (see \secref{subsec:human-behavioral-prediction}), and potentially  improves the efficiency of learning by reducing sample complexity. 
    
%\vspace{-0.05cm}
\section{Acknowledgements}
\label{sec:ack}
This work was funded in part by the Singapore Ministry of Education (grant MOE2016-T2-2-068), the National University of Singapore (grant R-252-000-587-112), US National Institute of Health R01 (grant R01EB019335), US National Science Foundation CPS (grant 1544797), US National Science Foundation NRI (grant 1637748), and the Office of Naval Research.

%% The file named.bst is a bibliography style file for BibTeX 0.99c
%\snnote{we are allowed only 1 extra page for references, so we may want to shorten some.}
\bibliographystyle{ACM-Reference-Format}
\bibliography{references}

\end{document}

%% file: hri-macros.tex
\usepackage{xspace}
\usepackage{tikz}
\usetikzlibrary{decorations.markings}

\newcommand{\human}{\textrm{H}\xspace}
\newcommand{\robot}{\textrm{R}\xspace}

\newcommand{\action} [2]{\ensuremath{a^{#1}_{#2}}}
\newcommand{\Action} [2]{\ensuremath{A^{#1}_{#2}}}

\newcommand{\ah}[1]{\action{\human}{#1}}
\newcommand{\ar}[1]{\action{\robot}{#1}}
\newcommand{\AH} {\Action{\human}{}}
\newcommand{\AR} {\Action{\robot}{}}
\newcommand{\aht}{\ah{t}\xspace}
\newcommand{\art}{\ar{t}\xspace}

\newcommand{\x}[1]{\ensuremath{x_{#1}}}
\newcommand{\X}{\ensuremath{X}}

\newcommand{\rwd}[1]{\ensuremath{r_{#1}}}

\newcommand{\policy} [1]{\ensuremath{\pi^{#1}}}
\newcommand{\policyh}{\policy{\human}\xspace}
\newcommand{\policyr}{\policy{\robot}\xspace}
\newcommand{\policyropt}{\ensuremath{\policy{\robot}_{*}}\xspace}

\newcommand{\trust}[1]{\ensuremath{\theta_{#1}}}

\newcommand{\perf}[1]{\ensuremath{e_{#1}}}

\newcommand{\baseline}{myopic\xspace}
\newcommand{\Baseline}{Myopic\xspace}

\newcommand{\mysquare}[1]{\tikz{\filldraw[draw=#1,fill=#1] (0,0)
rectangle (0.2cm,0.2cm);}}
\newcommand{\mycircle}[1]{\tikz{\filldraw[draw=#1,fill=#1] (0,0) circle [radius=0.1cm];}}
\newcommand{\mytriangle}[1]{\tikz{\filldraw[draw=#1,fill=#1] (0,0) --
(0.2cm,0) -- (0.1cm,0.2cm);}}

\tikzset{->-/.style={decoration={
  markings,
  mark=at position 0.5 with {\arrow{#1}}},postaction={decorate}}}
\newcommand{\myarrow}[1]{\tikz{\draw[->-={stealth}, line width=0.4mm, dashed] (0,0) to (0.6cm, 0cm);}}

%% file: figs/trust_model.tex
\begin{tikzpicture}[align=center]

% time t
‎\node [s] 				(x) {\x{t}};‎
\node [a, above right=0.5cm of x](ar) {\ar{t}};
\node [a, above =of ar](ah) {\ah{t}};
\node [s, above=3cm of x](theta) {\trust{t}};
\node [s, below=of theta] (perf) {\perf{t}};‎
% \node [r, below right=0.3cm of x] (r) {\rwd{t}};‎

%\path [draw, -X] (x) -- (ar);
\path [draw, -X] (x) -- (ah);
\path [draw, -X] (theta) -- (ah);
\path [draw, -X] (ar) -- (ah);
\path [draw, -X] (perf) -- (theta);
\path [draw, -X] (x) -- (perf);
% \path [draw, -X] (x) -- (r);

% time t+1
‎\node [s, right=2cm of x] 				(xn) {\x{t+1}};‎
\node [s, above=3cm of xn](thetan) {\trust{t+1}};
\node [s, below=of thetan] (perfn) {\perf{t+1}};‎
% \node [r, below right=0.3cm of xn] (rn) {\rwd{t+1}};‎

‎
\path [draw, -X] (x) -- (xn);
\path [draw, -X] (x) edge [bend right] (perfn);
\path [draw, -X] (theta) -- (thetan);
\path [draw, -X] (xn) -- (perfn);
\path [draw, -X] (ah) -- (perfn);
\path [draw, -X] (ar) -- (perfn);
\path [draw, -X] (perfn) -- (thetan);
\path [draw, -X] (ah) -- (xn);
\path [draw, -X] (ar) -- (xn);
% \path [draw, -X] (xn) -- (rn);

\end{tikzpicture}

%% file: figs/trust_flowchart.tex
\begin{tikzpicture}[align=center]

\node [b] (robot) {Robot};‎
\node [b, below=of robot, yshift=10pt] (human) {Human};‎
\node [b, below=of human, yshift=-12pt] (env) {Environment};‎
\begin{scope}[on background layer]
% \node [b, fit=(robot)(human), fill=black!10!white, inner sep=5pt, label=Team] (team) {};
\node [b, fit=(robot)(human), fill=black!10!white, inner xsep=5pt, inner ysep=16pt, label=Team] (team) {};
\end{scope}

\path [draw, -X] (robot) -- node [anchor=center, auto] {\ar{t}} (human);
% \path [draw, -X] (human) -- node [anchor=north, auto] {\ah{t}} (team);
% \path [draw, -X] (human) -- node [anchor=center, auto] {\ah{t}} (robot);
% \path [draw, -X] (robot) -- node [anchor=south east, auto] {\ar{t}} (human);
% \path [draw, -X] (human) -- node [anchor=north west, auto] {\ah{t}} (robot);
% \path [draw, -X] (team) -- node [anchor=center] 
% {\ah{t} \ar{t}} (env);
\path [draw, -X] (team) -- node [anchor=center, pos=0.5] 
{\ah{t} \ar{t}} (env);

% min edits
\draw [o=2pt] (human) -- node [anchor=east, pos=0.5] 
{\ah{t}} (team);

\coordinate [left=of env] (temp);
\path [draw, -X] (env) -- node [anchor=south, text width=1cm] 
{\x{t+1}} (temp) |- (team);
% \path [draw, -X] (env) -- node [anchor=center, text width=1cm] 
% {\x{t+1} \rwd{t+1}} (temp) |- (team);
\end{tikzpicture}